\documentclass[10pt,twocolumn,letterpaper]{article}

\usepackage[pagenumbers]{cvpr}

\usepackage[dvipsnames]{xcolor}
\usepackage{graphicx} 
\usepackage{multirow}
\usepackage{subcaption}
\usepackage{amsmath}
\usepackage{pgfplots}
\pgfplotsset{compat=1.18}
\usepackage{bm}
\usepackage{pifont}
\usepackage{adjustbox}
\usepackage{multirow, makecell}
\usetikzlibrary{intersections}
\usepgfplotslibrary{fillbetween}

\newcommand\norm[1]{\left\lVert#1\right\rVert}
\newcommand{\R}{\mathbb R}

\usepackage{color}

\definecolor{plot0}{HTML}{2282f5}
\definecolor{plot1}{HTML}{c56338}
\definecolor{plot2}{HTML}{58bf7f}
\definecolor{plot3}{HTML}{00338d}
\definecolor{plot4}{HTML}{f9d058}
\definecolor{plot5}{HTML}{ac61af}
\definecolor{plot6}{HTML}{98ab45}
\definecolor{plot7}{HTML}{bb464a}
\definecolor{plot8}{HTML}{2a4e00}
\definecolor{plot9}{HTML}{bf8a3a}

\definecolor{alpha0}{HTML}{b3d5ff}
\definecolor{alpha1}{HTML}{80b9ff}
\definecolor{alpha2}{HTML}{4d9dff}
\definecolor{alpha3}{HTML}{1981ff}
\definecolor{alpha4}{HTML}{005ccc}

\definecolor{beta0}{HTML}{ffad8c}
\definecolor{beta1}{HTML}{ff7f4d}
\definecolor{beta2}{HTML}{ff4800}

\usepackage{url}
\usepackage{float}

\newcommand{\mypara}[1]{\vspace{-2.5mm}\paragraph*{#1}}

\usepackage[accsupp]{axessibility} 

\definecolor{cvprblue}{rgb}{0.21,0.49,0.74}
\usepackage[pagebackref,breaklinks,colorlinks,citecolor=cvprblue]{hyperref}

\def\paperID{15170}
\def\confName{CVPR}
\def\confYear{2024}

\title{Back to 3D: Few-Shot 3D Keypoint Detection with Back-Projected 2D Features}

\author{Thomas Wimmer\textsuperscript{1,2} \qquad Peter Wonka\textsuperscript{3} \qquad Maks Ovsjanikov\textsuperscript{1}\\
{\textsuperscript{1}LIX, École Polytechnique \quad
\textsuperscript{2}Technical University of Munich \quad
\textsuperscript{3}KAUST}\\
{\tt\small thomas.m.wimmer@tum.de, pwonka@gmail.com, maks@lix.polytechnique.fr}
}

\begin{document}

\twocolumn[{
\maketitle
\begin{center}
    \captionsetup{type=figure}
    \centering
    \vspace{-5mm}
    \includegraphics[width=.99\textwidth]{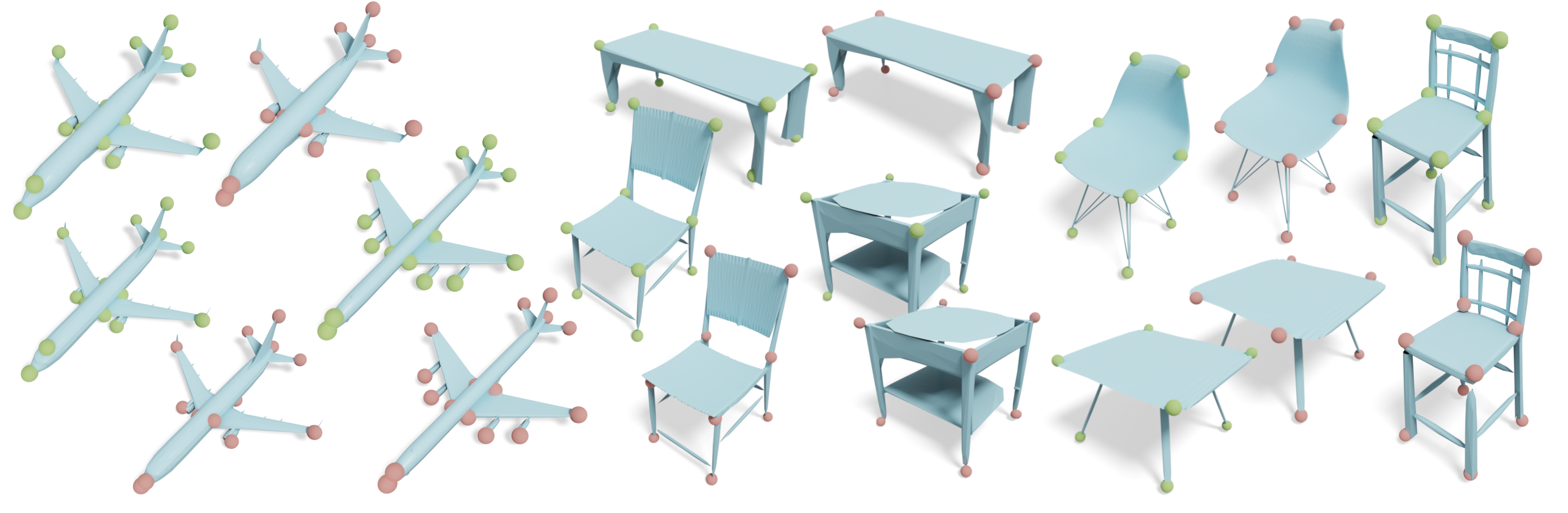}
    \vspace{-3mm}
    \captionof{figure}{Qualitative results of our proposed method B2-3D for few-shot keypoint detection using back-projected features (red) with ground truth keypoint annotations (green).}
    \label{fig:dino-keypoints}
\end{center}
}]

\begin{abstract}
With the immense growth of dataset sizes and computing resources in recent years, so-called foundation models have become popular in NLP and vision tasks.
In this work, we propose to explore foundation models for the task of keypoint detection on 3D shapes. A unique characteristic of keypoint detection is that it requires semantic and geometric awareness while demanding high localization accuracy.
To address this problem, we propose, first, to back-project features from large pre-trained 2D vision models onto 3D shapes and employ them for this task.
We show that we obtain robust 3D features that contain rich semantic information and analyze multiple candidate features stemming from different 2D foundation models.
Second, we employ a keypoint candidate optimization module which aims to match the average observed distribution of keypoints on the shape and is guided by the back-projected features. The resulting approach achieves a new state of the art for few-shot keypoint detection on the KeyPointNet dataset, almost doubling the performance of the previous best methods.
\end{abstract}

\section{Introduction}
Foundation models are finding their way into an increasing number of downstream applications.
They show strong generalization to tasks they were not explicitly trained for and exhibit surprising zero- or few-shot capabilities.
Successful approaches have been presented for processing text or 2D images~\citep{brown2020language, radford2021learning, oquab2023dinov2}, but there are no prominent 3D methods yet available that are aimed at \textit{local} details.
The main reasons for this are the diversity of 3D representations (i.e.,~meshes, point clouds, volumetric or implicit representations) and the comparably low availability of high-quality 3D data.

Recent works using 2D foundation models for shape analysis, like \citep{hegde2023clip, liu2023partslip, zhang2022pointclip}, have focused mainly on global tasks like shape classification, or segmentation which represents a middle ground between global and local analyses~\citep{abdelreheem2023satr}. This paper builds on this continuum, advancing from classification to segmentation and now to keypoints, each stage requiring a more localized understanding of geometric details.

In this work, we focus on the task of few-shot keypoint detection on 3D meshes. Repeatable keypoints enable a wide range of downstream applications, including rigid and non-rigid shape matching, object tracking, shape reconstruction, and shape manipulation \cite{bueno2016detection,wang2018learning,ovsjanikov2010one,chen2021unsupervised,jakab2021keypointdeformer} to name a few. Furthermore, the complex nature of the keypoint detection problem, which requires both semantic and geometric shape understanding, has been used in the past as a fruitful testbed for both exploiting and evaluating various local shape descriptors \cite{salti2015learning,boyer2011shrec} as well as consistency relations between 3D shapes and their views \cite{you2020keypointnet}.

A natural approach for keypoint detection is to leverage dense 3D geometric features or descriptors. Ideally, such features should capture both global (semantic) and local (geometric) information at the same time.
Previous works in this domain either focus on axiomatic features or propose to pre-train models on 3D datasets~\citep{bai2020d3feat,attaiki2022ncp}. Such methods, however, are limited by both robustness issues and the lack of diversity and amount of 3D training data. As a result, existing approaches tend to have limited accuracy on complex 3D models.
In contrast, we propose to retrieve informative features from powerful \textit{pre-trained} 2D vision encoders that can be used directly without any further training.

We show, for the first time, that such back-projected features enable a localized understanding of shape geometry while at the same time being able to capture rich global semantic information.
Specifically, we demonstrate that these features are robust against rotations and scaling changes and can be computed for any point on the shape's surface and with any texture type. Through assembling information from multiple rendered views around the shape, we show that even coarse 2D features can lead to dense 3D descriptors that vary continuously on the surface.

In addition to the lack of informative features, a common challenge in keypoint detection is the presence of symmetries. A successful method should detect \textit{all} keypoints across various symmetries (e.g.,~different legs of a table) while avoiding superfluous detections. To address this issue, we introduce a simple yet effective optimization strategy that leverages the back-projected features and ensures that the detected keypoints have a similar distribution on the target shapes as on the given few-shot examples.

Our main contributions are (1) the formalization of feature back-projection, including Gaussian geodesic re-weighting for handling noisy point visibility information, (2) the analysis of back-projected features from different recent foundation models and their properties, and (3) an optimization module which aims to match the average observed distribution of keypoints on the target shape. Our resulting method achieves state-of-the-art results on the KeypointNet benchmark, improving over the best-competing method by over \textbf{93\%} IoU on average over all evaluation distances. We compare different 2D feature extractors as well as axiomatic 3D descriptors and evaluate further components of our optimization module in an extensive ablation study. As an additional validation, we demonstrate the strong generalization of the features to other tasks and achieve state-of-the-art results in the task of part segmentation transfer.

\section{Related Work}
The flexibility of the transformer architecture and the increase in data and computing resources have led to the creation of powerful foundation models. This section gives an overview of relevant models and their transfer for the analysis of 3D data, as well as an overview of classic shape descriptors and previous methods for keypoint detection on 3D shapes.

\mypara{Foundation Models}
Large pre-trained models that were trained on vast quantities of data and exhibit strong generalization to new tasks with no or only little fine-tuning are referred to as foundation models.
Prominent examples include large language models that exhibit strong zero-shot generalization through prompting.
The success in the textual domain, as well as the high availability of image data on the internet, has led to recent advances in vision and multi-modal models.
Models like DINO~\citep{caron2021emerging,oquab2023dinov2} or masked autoencoders~\citep{he2022masked} are, in their essence, just feature extractors for image data as any other neural network.
However, their self-supervised training on large datasets has led to a great semantic understanding of scenes that can be utilized using simple linear or nearest-neighbor-based models to solve downstream tasks.
Multi-modal foundation models like the vision-language model (VLM) CLIP~\citep{radford2021learning} consist of separate encoder models that aim to map data from different modalities, like text and images, to the same meaningful embedding space.
By comparing the computed embeddings of candidate text prompts with an image embedding, one can thus perform, e.g., open-vocabulary zero-shot classification of images.
Finally, \citet{kirillov2023segment} proposed SAM, a foundation model specifically aimed at performing various segmentation tasks.

\mypara{Lifting Knowledge from 2D to 3D}
Several works have aimed at replicating the success of self-supervised pre-training in the 3D domain~\citep{pang2022masked, zhang2022point}.
However, these approaches are limited by the lack of high-quality training data in 3D.
Other methods thus try to transfer the meaningful embeddings from pre-trained 2D models like CLIP to 3D.
Several works focus on multi-modal contrastive pre-training, where they train a new 3D model that should be aligned with the embeddings computed from the frozen 2D CLIP encoder~\citep{gao2023ulip, zeng2023clip2, hegde2023clip, rozenberszki2022language, peng2023openscene}.
While these methods could exhibit similar zero-shot properties as VLMs, for example, for shape classification, they are limited by the lack of quantity and variety of 3D data for training.
Another approach is to bypass the training of 3D-specific models and apply 2D encoders directly to rendered views of the 3D object or scenes.
Several works propose to render shapes from different viewpoints, process the views with 2D CLIP encoders, average the resulting image embeddings over all views, and directly compare them with text prompt embeddings for tasks like zero-shot shape recognition~\citep{zhang2022pointclip, liu2023partslip, hegde2023clip}.
Other recent works propose to use pre-trained 2D segmentation or object detection models for shape part segmentation~\citep{abdelreheem2023satr} or subsequently for shape matching~\citep{abdelreheem2023zero}.
Parallel to this work, \citet{morreale2023neural} proposed to use pre-trained 2D feature extractors to obtain fuzzy matches between shapes that are then projected onto the 3D shape and subsequently refined.
While their approach is the most similar to our work, the core difference is that they propose to process the features in the 2D domain to obtain semantic correspondences and project these back onto the shapes, while we propose to back-project and aggregate the meaningful extracted features to the 3D shapes.
Our approach is thus more versatile and allows for various downstream tasks instead of solely shape matching.
A third way of using pre-trained VLMs is to optimize a separate network on a single instance, guided by the CLIP model (with frozen weights), while using differentiable rendering.
This approach has been used for shape stylization~\citep{michel2022text2mesh}, localization of semantic regions on shapes~\citep{decatur20233d} or mesh deformation~\citep{gao2023textdeformer}.

We render the 3D shape from multiple views and project features back from 2D encoders onto the surface.
As we demonstrate below, by doing so, we obtain high-quality pointwise feature descriptors that enable tasks that require high accuracy, like keypoint detection.

\mypara{3D Shape Descriptors}
A significant amount of hand-crafted shape descriptors is based on spectral methods, utilizing information from Laplacian eigenvalues and eigenfunctions of the shape.
The Laplace-Beltrami operator, as one of the fundamental tools in geometry processing, gave rise to early, well-established shape descriptors like HKS~\citep{sun2009concise} and WKS~\citep{aubry2011wave}.
Another class of hand-crafted features is based on local reference frames (LRF). A prominent example is the SHOT descriptor~\citep{salti2014shot}.
While spectral descriptors are often used for shape matching, LRF-based methods are mainly employed for 3D recognition or registration tasks.
With the rise of learning-based methods, recent attempts aim at using neural networks to learn shape descriptors~\citep{bai2020d3feat}.
While all neural methods naturally provide embeddings that can be extracted from their hidden layers, the difficulties have already been described before; as 3D data exists in various formats and there are only relatively small datasets available, such shape descriptors are usually limited in their quality and steered towards specific tasks, such as shape matching.

A general weakness of common shape descriptors is their dependence on mesh triangulation and quality (e.g., to compute robust Laplacian information), as well as missing semantic understanding.
Our proposed back-projected features can be computed for any point on the surface of a mesh. They provide high-quality semantic information and can include texture information if available while being robust to low mesh quality, surface holes, and other common problems for classic shape descriptors.

\mypara{3D Keypoint Detection}

The choice of method for 3D keypoint detection often depends on the desired downstream use. Classic keypoint detection methods are often hand-engineered methods that mark a (usually high number) of geometrically salient points on a shape regardless of their semantic meaning. In a subsequent step, these points can then be, e.g., matched with a scan at a different point in time and thus used for tasks such as shape matching or registration. \citet{tombari2013performance} provides an overview over classic 3D keypoint detection methods.

A more challenging task is that of finding a specific set of keypoints on 3D shapes. KeypointNet~\citep{you2020keypointnet} is a subset of the ShapeNet~\citep{shapenet2015} dataset with annotated keypoints for 16 shape categories, where for each model, there are between five and 23 semantic keypoints.
One can employ plain 3D models, such as PointNet~\citep{qi2017pointnet}, to detect keypoints from shapes in a supervised manner.
However, such approaches usually require large amounts of training data, which implies high costs for capturing and labeling 3D models when trying to apply these techniques in practice.

For these reasons, recent works proposed to instead focus on few-shot keypoint detection.
Approaches range from learning self-supervised 3D features and fitting custom detection modules on them~\citep{bai2020d3feat,attaiki2022ncp} to training unsupervised keypoint detection models on large unlabeled datasets followed by a few-shot selection of the detected keypoints~\citep{you2022ukpgan}.

\section{Method}
As mentioned above, we consider the few-shot keypoint detection problem. Thus, we are given a small set of shapes with known keypoints, and our aim is to detect keypoints on some new target shape. For this, our overall strategy consists of transferring the keypoints from the labeled (source) shapes onto the unlabeled (target) one. Our solution comprises two main components: a point similarity component which measures the similarity between vertices of the source and target shapes, and an optimization block which aims to preserve the overall distribution of keypoints and prevent collapses, e.g., due to symmetries. Crucially, for point similarity, we propose to back-project features given by 2D foundation models onto the 3D shapes. Below, we describe first our feature extraction strategy (Sec.~\ref{sec:feature-extraction}) and then the optimization module (Sec.~\ref{sec:optimization}) before presenting the analysis of computed features and results in the following sections. We call our method B2-3D.

\subsection{3D Feature Extraction} \label{sec:feature-extraction}

\begin{figure}[htb]
    \centering
    \hspace*{\fill}%
    \begin{subfigure}[b]{.28\columnwidth}
        \centering
        \raisebox{-0.5\height}{\includegraphics[width=.85\textwidth]{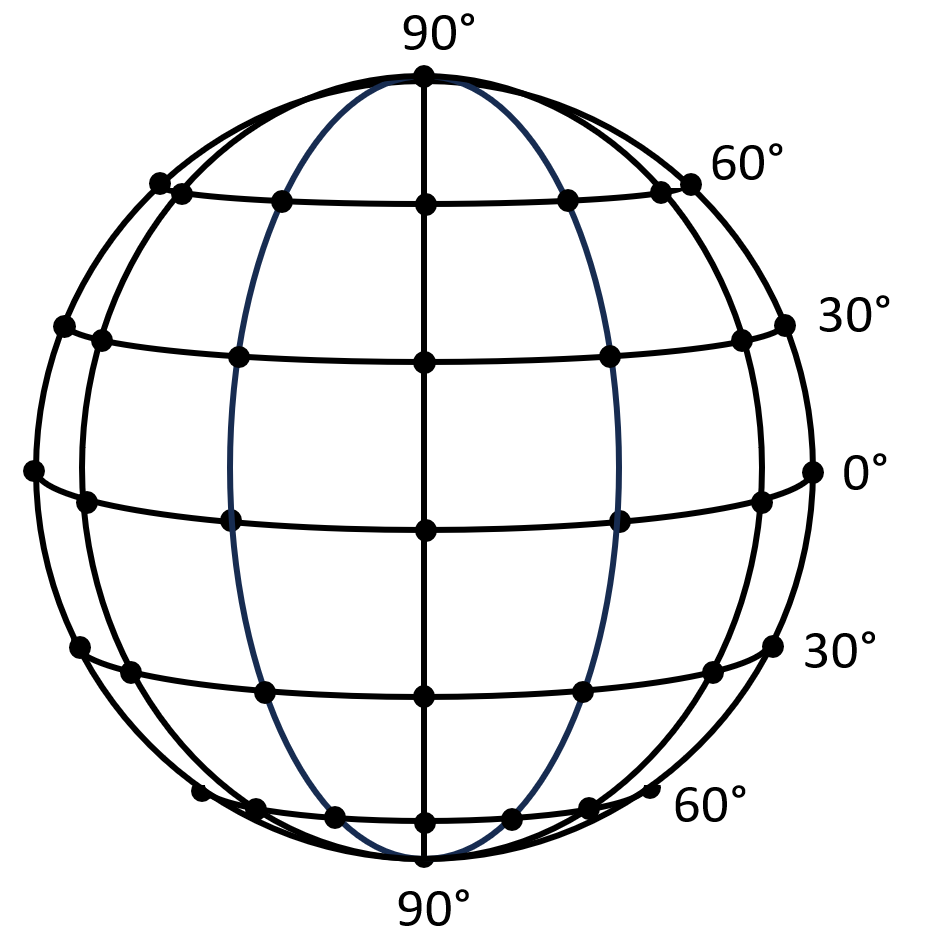}}
        \caption{View points used for rendering.}
        \label{fig:render-viewpoints}
    \end{subfigure}
    \hspace*{\fill}%
    \begin{subfigure}[b]{.6\columnwidth}
        \centering
        \raisebox{-0.5\height}{\includegraphics[width=\textwidth]{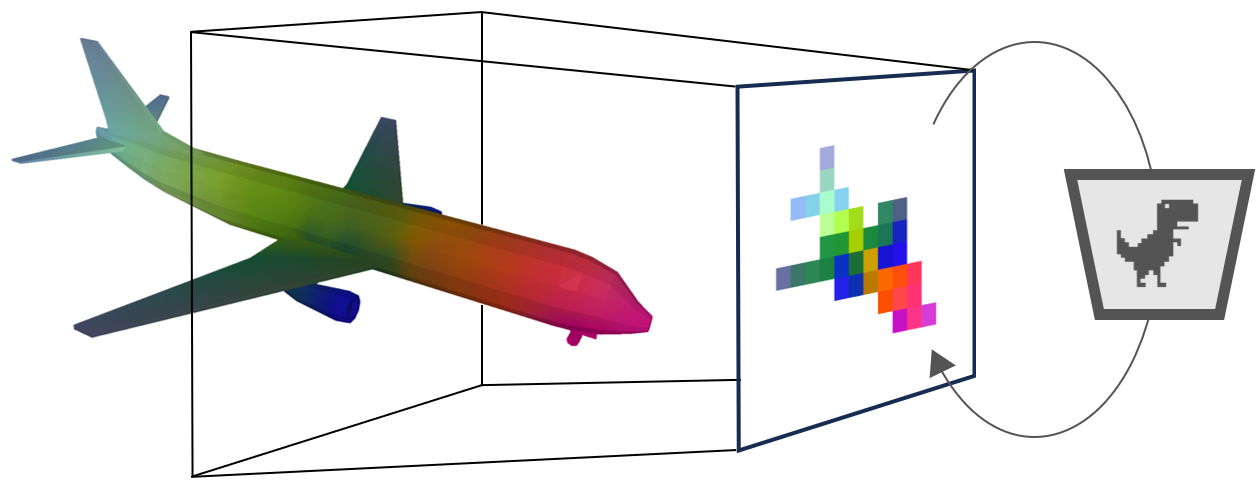}}
        \caption{Back-projection of features computed from rendered views.}
        \label{fig:dino-back-projection}
    \end{subfigure}
    \hspace*{\fill}%
    \caption{After processing the rendered views of an object with a large pre-trained vision encoder (e.g.,~DINO), we back-project the features onto the 3D shape and aggregate the information from all views (a) to obtain rich semantic 3D features (b).}
    \label{fig:back-projection}
\end{figure}

Our proposed pipeline for feature extraction consists of first rendering the object from multiple viewpoints around the object, processing the rendered views with a pre-trained 2D encoder, and back-projecting the computed features onto the 3D shape (see Fig.~\ref{fig:back-projection}). While we propose to use the DINO model~\citep{oquab2023dinov2}, we investigate different 2D feature extractors for shape analysis tasks in our experiments.

As we know the intrinsic camera parameters $K$, as well as the exact displacement $T$ and rotation $R$ of the cameras for all rendered views, we can compute the exact pixel location $(x,y)$ of any 3D point ($X_0$ in homogeneous coordinates) in the rendered images as
\begin{equation}
    \lambda (x,y,1)^T = K C_0 \bm{g} X_0 \quad \text{with} \quad \bm{g} =
    \begin{pmatrix}
        R & T\\
        0 & 1
    \end{pmatrix},
\end{equation}
where we use a normal perspective camera with the standard projection matrix $C_0$ and the scalar factor $\lambda$.

Using PyTorch3D~\citep{ravi2020pytorch3d} for rendering, we can additionally determine whether a 3D point is visible in a rendered image. If it is, we assign the feature at the corresponding computed pixel location $(x,y)$ to it.
To aggregate the back-projected features from all views, we simply average them.

When projecting features only to the points visible in the rendered views, we observe that for more complex meshes, the point visibility information is not of sufficient quality to get a noise-free signal.
To bypass this problem, we propose Gaussian geodesic re-weighting of the back-projected features which is essentially a Gaussian smoothing of the features along the surface.
The feature $f_i$ for a point $i$ on the surface can then be computed as
\begin{equation}
\label{eq:gaussian-reweighting}
\begin{split}
    f_i &= \cfrac{1}{\sum_j w(d_{ij})} \sum_{j} w(d_{ij}) f^{(b)}_j,\\
    \text{ with }w(d_{ij}) &= \exp(-d_{ij}^2 / 2\sigma^2),
\end{split}
\end{equation}
using the geodesic distance information $d_{i,j}$ between pairs of points $(i,j)$, the back-projected features for visible points $f^{(b)}$, and a standard deviation $\sigma$ that is left as a hyperparameter.

\subsection{Few-Shot Keypoint Detection} \label{sec:optimization}
Our goal is to capitalize on the rich semantic features won through back-projection for the task of keypoint detection.  
We aim at doing so in a few-shot setting, where we are given only a few shapes for a given class with an annotated set of keypoints and predict corresponding keypoint locations on a new shape from the same class.

When simply matching features of given keypoints with features for candidate positions on a new shape, the results for symmetric keypoints (e.g., the ends of table legs) may suffer from finding only a subset of the relevant points (e.g., all keypoints get matched to the same table leg on the new shape).
To properly handle such symmetries, we propose an optimization of the keypoint locations that aims to match the global distribution of keypoints on the new shape to prevent a collapse.

Given the features $F_{\text{kp}} \in \R^{k \times d_\text{emb}}$ computed for $k$ keypoints and the pairwise relative geodesic distances between these keypoints $D_{\text{kp}} \in [0,1]^{k \times k}$ on a given shape, we can find the corresponding keypoints on a new shape with $n$ candidate locations, for which we compute the features $F_{\text{cand}} \in \R^{n \times d_\text{emb}}$ and the pairwise relative geodesic distances $D_{\text{cand}} \in [0,1]^{n \times n}$.
Usually, the dimensions are $k \approx 10$ and $n = 2048$ in our experiments, where the keypoint candidate locations can be simply obtained through farthest-point sampling from the surface.

We can now define an optimization to find the best locations among the $n$ keypoint candidates on the new shape.
To do so, we define a right-stochastic selection matrix ${S \in [0,1]^{n \times (k + 1)}}$, where for each candidate point $i$, the probability that this point corresponds to the keypoint $j$ is given by the value $S_{ij}$.
The last, additional column shows the probability of a point not corresponding to any of the keypoints. We impose the right-stochastic character on $S$ by applying a softmax per row at every optimization step.
For convenience, we define ${\hat{S} \in [0,1]^{n \times k} = S_{[1,..,n; 1,..,k]}}$ as the first $k$ columns of the matrix $S$.
We then formulate the optimization objective $L = L_{\text{feature}} + \alpha L_{\text{distance}}$ with
\begin{align}
\begin{split}
    L_{\text{feature}} &= \norm{\hat{S}^T F_{\text{cand}} - F_{\text{kp}}}_2,\\
    L_{\text{distance}} &= \norm{\hat{S}^T D_{\text{cand}} \hat{S} - D_{\text{kp}}}_2.
\end{split}
\end{align}

To extract the corresponding keypoints, we simply take the argmax over the columns of $\hat{S}$ after our optimization has finished.

By formalizing the keypoint search as an optimization problem, we can integrate objectives like matching the relative geodesic distances between keypoints, besides simply matching the computed features of given keypoints with the candidate points.
By doing so, we also move from a per-keypoint solution to a global optimization over all keypoints at the same time, which we show to be useful in our experiments.
Note that our proposed keypoint optimization module is agnostic to the used shape descriptor. We ablate the choice of DINO features in our experiments (Sec.~\ref{sec:shape-descriptors}).

We normalize the geodesic distances between different points to $[0,1]$ by division through the maximal observed geodesic distance on the given shape.
The use of such pairwise relative geodesic distances is more robust than simply matching average 3D positions of observed key points or non-normalized geodesic distances, which are sensitive to shape alignment and are not rotation- or scale-invariant.

When given multiple labeled samples, one can simply average the distance matrices and features per keypoint class over all samples. Additionally, we experiment with the use of a retrieval module, where we first retrieve the closest match in the labeled shapes and then only use the distance and keypoint information from the retrieved shape. For retrieval, we additionally use the class token of the DINO model, average it over all viewpoints, and perform a nearest-neighbor search in the feature space to find the best match with the unseen shape. We analyze the performance of this retrieval module in our experiments.

\section{Analysis of Back-Projected Features}

Before diving deeper into the task of keypoint detection, we investigate several important properties of back-projected features. In this section, we focus on the features back-projected from the DINO model~\citep{oquab2023dinov2}, which produces the best results in our keypoint detection experiments (Sec. \ref{sec:experiments}).

\subsection{Feature Stability} \label{sec:feature-stability}
We first analyze how stable the found elements behave under changes in certain parameters of the extraction process.
This analysis is of great importance as one of the key limitations of the rendering- and back-projection-based method is its sensitivity to the quality of rendered images and other scene and capture parameters.

\begin{figure}[htb]
    \centering
    \resizebox{\columnwidth}{!}{
         \begin{tikzpicture}

    \begin{axis}[%
    name=plot1,
    ymin=0.7, ymax=1.001,
    xmin=-8, xmax=196,
    height=.33\textwidth,
    width=.3\textwidth,
    xtick={0, 50, 100, 150},
    ytick={0.9, 0.99, 1.0},
    ylabel near ticks,
    y coord trafo/.code={\pgfmathparse{ln((#1 - 0.002) / (1 - (#1 - 0.002)))}},
    y coord inv trafo/.code={\pgfmathparse{1 / (1 + exp(-(#1+0.002)))}\pgfmathresult},
    ymajorgrids=true,
    grid style=dashed,
    every axis plot/.append style={line width=1.5pt},
    xticklabel=\pgfkeys{/pgf/number format/.cd,fixed,precision=0,zerofill}\pgfmathprintnumber{\tick},
    yticklabel=\pgfkeys{/pgf/number format/.cd,fixed,precision=2,zerofill}\pgfmathprintnumber{\tick},
    xlabel style={align=center, at={(0.5,-0.08)}, anchor=north}, xlabel={ Number of views},
    ylabel style={align=center, at={(-0.22, 0.5)}, anchor=south}, ylabel={ Similarity to features\\[-2pt] with most views},
    ]
        \addplot[color=plot0] table [x=views, y=views_mean, col sep=comma] {numbers/stability-views.csv}; \label{stability-views}
        \addplot[name path=views_top, color=plot0, draw opacity=0] table [x=views, y=views_top, col sep=comma] {numbers/stability-views.csv};
        \addplot[name path=views_down, color=plot0, draw opacity=0] table [x=views, y=views_down, col sep=comma] {numbers/stability-views.csv};
        \addplot[plot0!50,fill opacity=0.4] fill between[of=views_top and views_down];
    \end{axis}
    \hspace{.01\textwidth}%
    \begin{axis}[%
    name=plot2,
    ymin=0.93, ymax=1.01,
    xmin=0.97, xmax=1.63,
    height=.33\textwidth,
    width=.3\textwidth,
    xtick={1.0, 1.2, 1.4, 1.6},
    ytick={0.94, 0.96, 0.98, 1.0},
    ylabel near ticks,
    ymajorgrids=true,
    grid style=dashed,
    every axis plot/.append style={line width=1.5pt},
    xticklabel=\pgfkeys{/pgf/number format/.cd,fixed,precision=1,zerofill}\pgfmathprintnumber{\tick},
    yticklabel=\pgfkeys{/pgf/number format/.cd,fixed,precision=2,zerofill}\pgfmathprintnumber{\tick},
    xlabel style={align=center, at={(0.5,-0.08)}, anchor=north}, xlabel={ Relative render distance},
    ylabel style={align=center, at={(-0.22, 0.5)}, anchor=south}, ylabel={ Similarity to features\\[-2pt] with smallest distance},
    at=(plot1.right of south east), anchor=left of south west,
    ]
        \addplot[color=plot0] table [x=dist, y=dist_mean, col sep=comma] {numbers/stability.csv}; \label{stability-distance}
        \addplot[name path=dist_top, color=plot0, draw opacity=0] table [x=dist, y=dist_top, col sep=comma] {numbers/stability.csv};
        \addplot[name path=dist_down, color=plot0, draw opacity=0] table [x=dist, y=dist_down, col sep=comma] {numbers/stability.csv};
        \addplot[plot0!50,fill opacity=0.4] fill between[of=dist_top and dist_down];
    \end{axis}
    \hspace*{.01\textwidth}%
    \begin{axis}[%
    name=plot3,
    ymin=0.989, ymax=1.001,
    xmin=-0.1, xmax=6.3,
    height=.33\textwidth,
    width=.3\textwidth,
    xtick={0,3.14,6.28},
    xticklabels={0, $\pi$, $2\pi$},
    ytick={0.99, 0.992, 0.994, 0.996, 0.998, 1.0},
    ylabel near ticks,
    ymajorgrids=true,
    grid style=dashed,
    every axis plot/.append style={line width=1.5pt},
    yticklabel=\pgfkeys{/pgf/number format/.cd,fixed,precision=3,zerofill}\pgfmathprintnumber{\tick},
    xlabel style={align=center, at={(0.5,-0.08)}, anchor=north}, xlabel={ Rotation},
    ylabel style={align=center, at={(-0.26, 0.5)}, anchor=south}, ylabel={ Similarity to features\\[-2pt] with no rotation},
    at=(plot2.right of south east), anchor=left of south west,
    ]
        \addplot[color=plot0] table [x=rot, y=rot_mean, col sep=comma] {numbers/stability.csv}; \label{stability-rot}
        \addplot[name path=rot_top, color=plot0, draw opacity=0] table [x=rot, y=rot_top, col sep=comma] {numbers/stability.csv};
        \addplot[name path=rot_down, color=plot0, draw opacity=0] table [x=rot, y=rot_down, col sep=comma] {numbers/stability.csv};
        \addplot[plot0!50,fill opacity=0.4] fill between[of=rot_top and rot_down];
    \end{axis}
\end{tikzpicture}
    }
    \caption{Feature stability analysis measuring the mean cosine similarity (with standard deviation in light blue) of extracted point features when applying modifications to the rendering process.}
    \label{fig:feature-stability}
\end{figure}
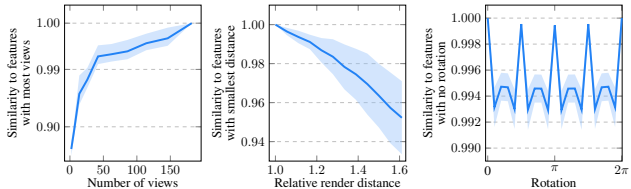

One of these crucial parameters is the number of different viewpoints from which we render the object.
In our experiments, we use a sampling strategy for the camera positions that partitions the unit sphere around the object in $n$ equidistant horizontal slices and spreads $2(n+1)$ viewpoints equiangular on the circumference of each slice, as well as one view from the top and one from the bottom of the object.
When varying this sample parameter $n$, we can observe that the features seem to converge at around a total number of 50 views (see Fig.~\ref{fig:feature-stability}). In our further experiments, we thus use $n=5$, resulting in 62 viewpoints around the object, as shown in Fig.~\ref{fig:render-viewpoints}.

Generally, we observe that although the features back-projected from a single 2D image are coarse, features assembled from multiple views around the object are smoother and with a higher detail level, as shown in Fig.~\ref{fig:dino-back-projection}.

\begin{figure}[htb]
    \centering
    \includegraphics[width=\columnwidth]{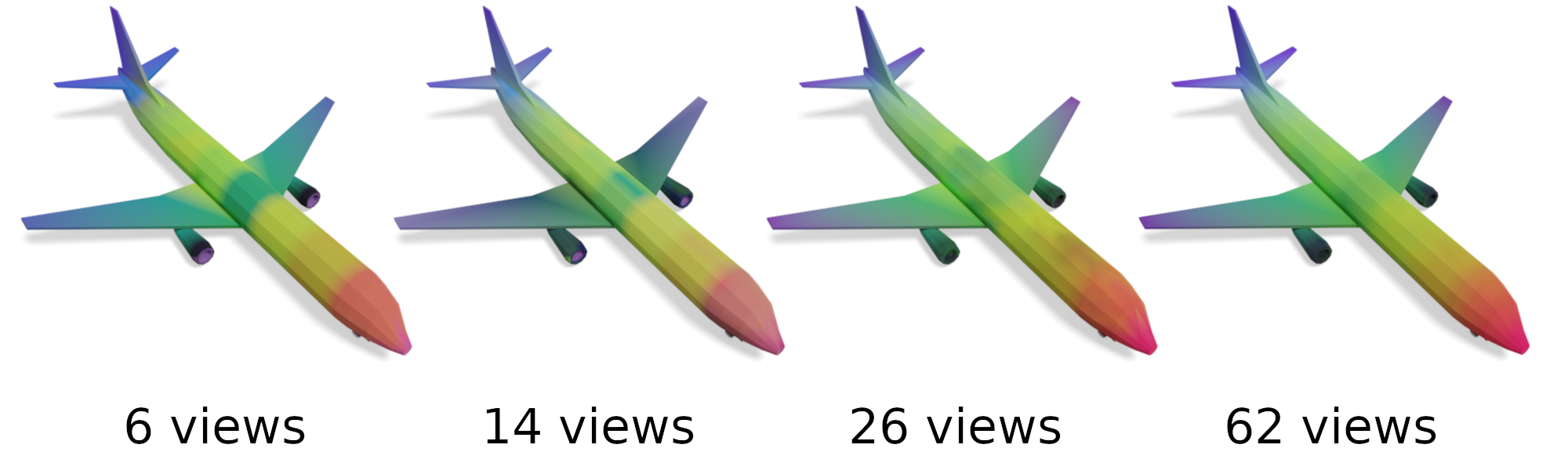}
    \caption{Increase in feature quality and distinctiveness with increasing number of rendering viewpoints. Visualization using a PCA, as described in Sec.~\ref{sec:semantic-analysis}.}
    \label{fig:features-with-k-views}
\end{figure}

We visualize the effect of increasing the number of rendering viewpoints in Fig.~\ref{fig:features-with-k-views}. As can be observed, the features get more distinctive and detailed with more rendered views, while with only a few views, one can clearly identify the patch-based architecture of the underlying vision transformer.

We further analyze the effects of changing the camera distance during rendering on the features. We find that there is an almost linear relationship between the increase in distance and the decrease in similarity to features captured at the minimal distance where all views capture the full object. The features are likely to collapse with too large render distances as more points of the shape are back-projected to the same patch in the 2D image. In practice, this does not pose a problem to our method, as shapes can be normalized to a certain scale in a pre-processing step.

For rendering, we employ PyTorch3D using one light source at the camera position for each rendering view.
Through doing so, we ensure an even lighting of the shape for feature extraction, as well as robustness to rotations around the up-axis of the object, which we assume as given. In our analysis, we observe that through the high number of viewpoints spread around the shape, the cosine similarity of features computed for rotated versions of shapes always stays above 0.99, thus proving the robustness of our capturing strategy.
The three peaks in similarity when rotating the object can be explained by the perfect alignment of rotation with the viewpoints, with rotations of 90, 180, and 270 degrees, respectively.

\subsection{Semantic Awareness} \label{sec:semantic-analysis}

\begin{figure}[htb]
    \centering
    \includegraphics[width=\columnwidth]{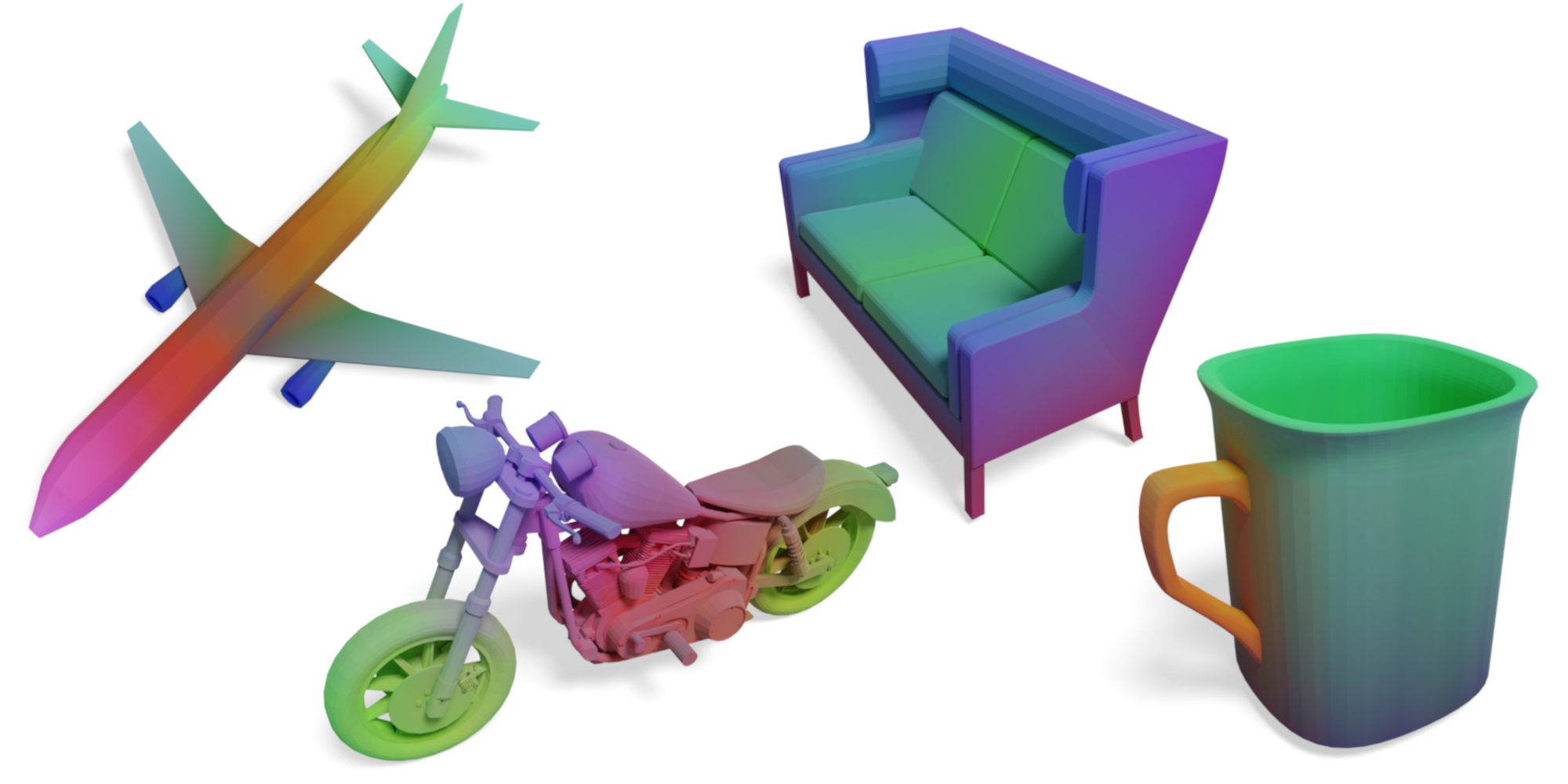}
    \caption{Back-projected ViT features on a shape can be visualized after performing a PCA to just three values per vertex. The extracted features contain rich semantic information and clearly assign different values to different semantic parts of the object.}
    \label{fig:dino-analysis-4}
\end{figure}

In order to visualize the retrieved features, we can compute them for all vertices of a given mesh and perform a dimensionality reduction using a PCA to get an interpretable 3-dimensional color vector for each point.
An example is shown in Fig.~\ref{fig:dino-analysis-4}.

The visualization of the principal components of the features suggests that the proposed method of back-projecting features on the 3D shape produces semantic features as similar, e.g., symmetric parts of the objects also get assigned similar values.
This visualization further highlights that through this simple back-projection technique, we are able to lift the powerful features from the 2D encoder to the three-dimensional shape.

\subsection{Geometric Properties}

Back-projected features comprise semantic information about scenes, which can be helpful for various downstream applications in shape analysis, as this is often a missing component of geometry-based methods.
However, an additional important property of shape descriptors is the understanding of pure \textit{geometric} information.

\begin{figure}[t!]
    \centering
    \includegraphics[width=.6\columnwidth]{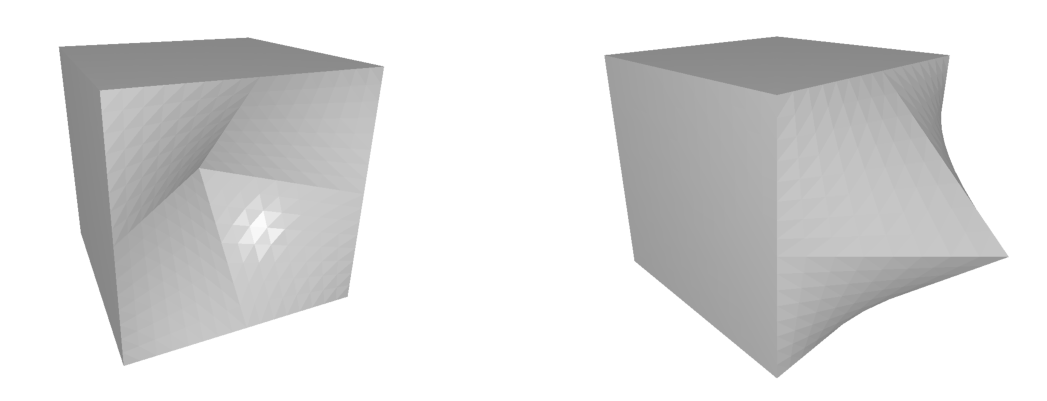}
    \includegraphics[width=.9\columnwidth]{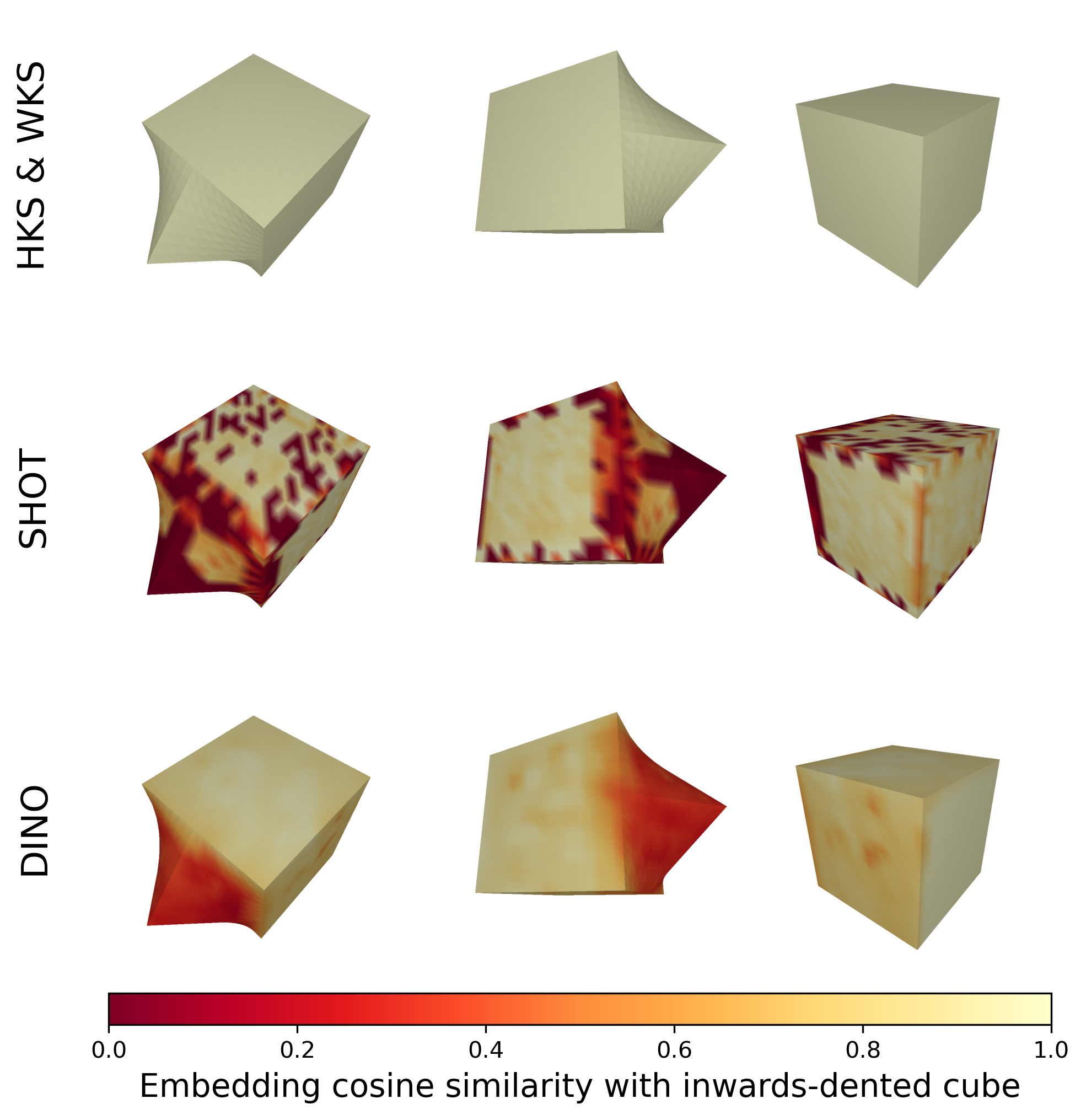}
    \caption{We compute different shape descriptors on two isometric cubes, once with an inward dent and once with an outward dent. Since the HKS and WKS signatures are based on geodesic information, the computed features for the different shapes do not change. The SHOT descriptor is sensitive to changes in triangulation, as well as to its support radius parameter, and results in noisy, changing features across the cube. Back-projected DINO features show significantly better performance as they change in the modified parts of the cube, while the features remain the same for the unmodified parts of the shape.}
    \label{fig:geodesic-boxes}
\end{figure}

To investigate these properties, we analyze the extracted features for two isometric cubes, where one has an inwards-dented side which is dented outwards on the other cube (see Fig.~\ref{fig:geodesic-boxes}). While the two shapes are isometric, they are obviously not the same, and intuitively, a good shape descriptor should reflect the local geometry change with a change in the respective local features.

Classic shape descriptors that are based on the shape Laplacian, like the HKS \cite{sun2009concise} and WKS \cite{aubry2011wave}, fail to distinguish the two cubes and assign the exact same features for all points on the shape, as the geodesic pairwise distances between any pair of points on the surface stay the same. The SHOT descriptor \cite{salti2014shot} is sensitive to triangulation changes and its support radius parameter, resulting in the noisy behavior shown in Fig.~\ref{fig:geodesic-boxes}. In contrast, our back-projected DINO features show a strong, localized reaction to the change in geometry while remaining similar in the unmodified parts of the shape.

Generally, we observe that the uptake to small local modifications of the shape is usually also local, with features for unrelated parts on the shape not being affected. Considering the global attention mechanism over the whole image in a ViT, this is an interesting and encouraging finding. Further examples of this behavior are shown in the supplementary materials. 

\section{Experiments}
\label{sec:experiments}
\subsection{Few-Shot Keypoint Detection}

\vspace{2mm}
\mypara{Setup}
We evaluate our method B2-3D on the KeypointNet dataset~\citep{you2020keypointnet}.
For each class, we select three random models from the KeypointNet dataset and use them as our few-shot samples.
We compare our results to several baseline methods for few-shot keypoint detection: SIFT-3D~\citep{rister2017volumetric}, HARRIS-3D~\citep{sipiran2011harris}, ISS~\citep{zhong2009intrinsic}, D3Feat~\citep{bai2020d3feat}, USIP~\citep{li2019usip}, UKPGAN~\citep{you2022ukpgan}, and FSKD~\citep{attaiki2022ncp}.

We use the same evaluation strategy as \citet{you2022ukpgan}, which computes the IoU of predicted and ground-truth keypoints from KeypointNet with a varying distance threshold for the evaluation.
An intersection is counted if the geodesic distance between a ground-truth keypoint and a predicted keypoint is smaller than this distance threshold.
We also stick to the same three classes of the dataset for evaluation: airplane, chair, and table.

\subsubsection{Quantitative and Qualitative Analysis}

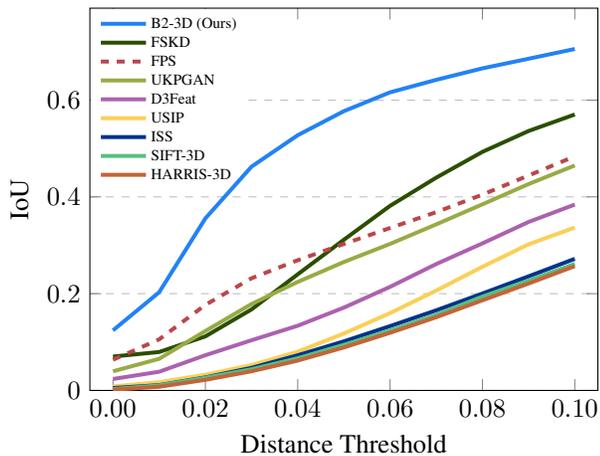
\begin{figure}[htb]
    \centering
    \begin{tikzpicture}
        \begin{axis}[
            width=\columnwidth,
            height=.8\columnwidth,
            xlabel={Distance Threshold},
            ylabel={IoU},
            xmin=-0.005, xmax=0.105,
            ymin=0, ymax=0.79,
            xtick={0, 0.02, 0.04, 0.06, 0.08, 0.1},
            xticklabel=\pgfkeys{/pgf/number format/.cd,fixed,precision=2,zerofill}\pgfmathprintnumber{\tick},
            ylabel near ticks,
            ymajorgrids=true,
            grid style=dashed,
            legend style = {at = {(0.001, 0.999)}, anchor = north west, draw = none, nodes={scale=0.58, transform shape}, fill opacity = 1.0, text opacity = 1},
            legend cell align={left}
            ]
            \addlegendimage{/pgfplots/refstyle=results-ours} \addlegendentry{B2-3D (Ours)}
            \addlegendimage{/pgfplots/refstyle=results-fskd} \addlegendentry{FSKD}
            \addlegendimage{/pgfplots/refstyle=results-fps} \addlegendentry{FPS}
            \addlegendimage{/pgfplots/refstyle=results-ukpgan} \addlegendentry{UKPGAN}
            \addlegendimage{/pgfplots/refstyle=results-d3feat} \addlegendentry{D3Feat}
            \addlegendimage{/pgfplots/refstyle=results-usip} \addlegendentry{USIP}
            \addlegendimage{/pgfplots/refstyle=results-iss} \addlegendentry{ISS}
            \addlegendimage{/pgfplots/refstyle=results-sift3d} \addlegendentry{SIFT-3D}
            \addlegendimage{/pgfplots/refstyle=results-harris3d} \addlegendentry{HARRIS-3D}
        \end{axis}
        \begin{axis}[
            width=\columnwidth,
            height=.8\columnwidth,
            xmin=-0.005, xmax=0.105,
            ymin=0, ymax=0.79,
            axis x line=none,
            axis y line=none,
            every axis plot/.append style={line width=1.5pt}
            ]

            \addplot[color=plot0] table [x=Threshold, y=Full, col sep=comma] {numbers/all.csv}; \label{results-ours}

            \addplot[color=plot8] table [x=Threshold, y=FSKD, col sep=comma] {numbers/all.csv}; \label{results-fskd}
            \addplot[color=plot7, dashed] table [x=Threshold, y=FPS, col sep=comma] {numbers/all.csv}; \label{results-fps};
            \addplot[color=plot6] table [x=Threshold, y=UKPGAN, col sep=comma] {numbers/all.csv}; \label{results-ukpgan};
            \addplot[color=plot5] table [x=Threshold, y=D3FEAT, col sep=comma] {numbers/all.csv}; \label{results-d3feat};
            \addplot[color=plot4] table [x=Threshold, y=USIP, col sep=comma] {numbers/all.csv}; \label{results-usip};
            \addplot[color=plot3] table [x=Threshold, y=ISS, col sep=comma] {numbers/all.csv}; \label{results-iss};
            \addplot[color=plot2] table [x=Threshold, y=SIFT-3D, col sep=comma] {numbers/all.csv}; \label{results-sift3d};
            \addplot[color=plot1] table [x=Threshold, y=HARRIS-3D, col sep=comma] {numbers/all.csv}; \label{results-harris3d};
        \end{axis}
    \end{tikzpicture}
    \caption{Our algorithm reaches a new state-of-the-art in few-shot keypoint detection by a large margin. 
    }
    \label{fig:few-shot-results}
\end{figure}

On KeypointNet, B2-3D outperforms the previous state-of-the-art by a very significant margin on all distance thresholds (Fig.~\ref{fig:few-shot-results}).
It reaches similar IoU levels as FSKD at a distance threshold of 0.1, already at a threshold of around 0.045.
The mean relative improvement over FSKD lies at 93\% with over 200\% improvement at a distance threshold of 0.02.
Qualitative results are shown in Figure~\ref{fig:dino-keypoints}.

We also include a comparison to farthest-point sampling with the average number of keypoints observed in the few-shot samples. Our implementation also uses the maximum of the average geodesic distance to seed the first point.
Surprisingly, this simple strategy, without any understanding of the underlying semantics or geometry, outperforms many of the previous works.
This can be explained by how we normally select keypoints: In order to have keypoints that describe the objects as well as possible, we tend to choose extreme points that are well distributed over the shape as keypoints. Nevertheless, such a simple baseline is insufficient for accurate keypoint detection.

\subsubsection{Ablation Studies}

\begin{figure}[htb]
    \centering
    \begin{tikzpicture}
    \begin{axis}[
        axis x line=none,
        axis y line=none,
        legend style = {at = {(0.985, 0.01)}, anchor = south east, draw = none, nodes={scale=0.6, transform shape}, fill opacity = 1.0, text opacity = 1},
        legend cell align={left}
        ] 
        \addplot[opacity=0,forget plot] {0};
        \addlegendimage{/pgfplots/refstyle=ablation-0} \addlegendentry{B2-3D}
        \addlegendimage{/pgfplots/refstyle=ablation-3} \addlegendentry{ViT-S}
        \addlegendimage{/pgfplots/refstyle=ablation-2} \addlegendentry{Texture}
        \addlegendimage{/pgfplots/refstyle=ablation-9} \addlegendentry{Retrieval}
        \addlegendimage{/pgfplots/refstyle=ablation-1} \addlegendentry{1-Shot}
        \addlegendimage{/pgfplots/refstyle=ablation-4} \addlegendentry{KNN}
        \addlegendimage{/pgfplots/refstyle=ablation-7} \addlegendentry{FSKD}
    \end{axis}
    \begin{axis}[
        width=\columnwidth,
        height=.8\columnwidth,
        xlabel={Distance Threshold},
        ylabel={Relative IoU},
        xmin=-0.005, xmax=0.105,
        ymin=0.25, ymax=1.08,
        xtick={0, 0.02, 0.04, 0.06, 0.08, 0.1},
        xticklabel=\pgfkeys{/pgf/number format/.cd,fixed,precision=2,zerofill}\pgfmathprintnumber{\tick},
        ylabel near ticks,
        ymajorgrids=true,
        grid style=dashed,
        every axis plot/.append style={line width=1.5pt}
    ]
        \addplot[color=plot1] table [x=Threshold, y=1-Shot, col sep=comma] {numbers/relative_ablation.csv}; \label{ablation-1}
        \addplot[name path=1-shot_top, color=plot1, draw opacity=0] table [x=Threshold, y=1-Shot-Up, col sep=comma] {numbers/relative_ablation.csv};
        \addplot[name path=1-shot_down, color=plot1, draw opacity=0] table [x=Threshold, y=1-Shot-Down, col sep=comma] {numbers/relative_ablation.csv};
        \addplot[plot1!50,fill opacity=0.4] fill between[of=1-shot_top and 1-shot_down];

        \addplot[color=plot9] table [x=Threshold, y=Retrieval, col sep=comma] {numbers/relative_ablation.csv}; \label{ablation-9}
        \addplot[name path=retrieval_top, color=plot9, draw opacity=0] table [x=Threshold, y=Retrieval-Up, col sep=comma] {numbers/relative_ablation.csv};
        \addplot[name path=retrieval_down, color=plot9, draw opacity=0] table [x=Threshold, y=Retrieval-Down, col sep=comma] {numbers/relative_ablation.csv};
        \addplot[plot9!50,fill opacity=0.4] fill between[of=retrieval_top and retrieval_down];
        
        \addplot[color=plot2] table [x=Threshold, y=Texture, col sep=comma] {numbers/relative_ablation.csv}; \label{ablation-2}
        \addplot[name path=texture_top, color=plot2, draw opacity=0] table [x=Threshold, y=Texture-Up, col sep=comma] {numbers/relative_ablation.csv};
        \addplot[name path=texture_down, color=plot2, draw opacity=0] table [x=Threshold, y=Texture-Down, col sep=comma] {numbers/relative_ablation.csv};
        \addplot[plot2!50,fill opacity=0.4] fill between[of=texture_top and texture_down];

        \addplot[color=plot3] table [x=Threshold, y=ViT-S, col sep=comma] {numbers/relative_ablation.csv}; \label{ablation-3}
        \addplot[name path=vit-s_top, color=plot3, draw opacity=0] table [x=Threshold, y=ViT-S-Up, col sep=comma] {numbers/relative_ablation.csv};
        \addplot[name path=vit-s_down, color=plot3, draw opacity=0] table [x=Threshold, y=ViT-S-Down, col sep=comma] {numbers/relative_ablation.csv};
        \addplot[plot3!50,fill opacity=0.4] fill between[of=vit-s_top and vit-s_down];
        
        \addplot[color=plot0] table [x=Threshold, y=Full, col sep=comma] {numbers/relative_ablation.csv}; \label{ablation-0}
        \addplot[name path=full_top, color=plot0, draw opacity=0] table [x=Threshold, y=Full-Up, col sep=comma] {numbers/relative_ablation.csv};
        \addplot[name path=full_down, color=plot0, draw opacity=0] table [x=Threshold, y=Full-Down, col sep=comma] {numbers/relative_ablation.csv};
        \addplot[plot0!50,fill opacity=0.4] fill between[of=full_top and full_down];

        
        

        \addplot[color=plot4] table [x=Threshold, y=KNN, col sep=comma] {numbers/relative_ablation.csv}; \label{ablation-4}
        \addplot[name path=knn_top, color=plot4, draw opacity=0] table [x=Threshold, y=KNN-Up, col sep=comma] {numbers/relative_ablation.csv};
        \addplot[name path=knn_down, color=plot4, draw opacity=0] table [x=Threshold, y=KNN-Down, col sep=comma] {numbers/relative_ablation.csv};
        \addplot[plot4!50,fill opacity=0.4] fill between[of=knn_top and knn_down];
        
        \addplot[color=plot8, dashed] table [x=Threshold, y=FSKD, col sep=comma] {numbers/relative_ablation.csv}; \label{ablation-7}

    \end{axis}
\end{tikzpicture}
    \caption{Relative performance (with standard deviation) compared to our optimization with features extracted from 3 shapes without textures with the DINO ViT-G model. We observe that features extracted from the smaller ViT result in almost equal performance, using (low quality) texture information slightly distracts the model, and the 1-shot performance has a higher variance but still always outperforms the previous state-of-the-art 3-shot method FSKD by a large margin. The reduced performance with simple nearest-neighbor-based keypoint detection [KNN] stresses the effectiveness of our optimization module in matching the keypoint distributions.}
    \label{fig:ablations}
\end{figure}
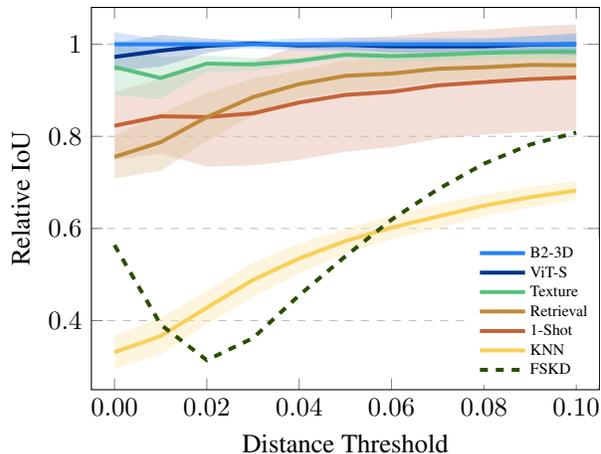

We examine the reasons for the success of our method with several ablation studies on different parts of the proposed pipeline. Our findings are illustrated in Fig.~\ref{fig:ablations}.

We find that changing the large DINO model for a distilled version [ViT-S] does not significantly harm the performance, but using the given textures of ShapeNet meshes does [Texture]. This is consistent with the observations made by \citet{morreale2023neural}; low-quality texture information is more likely to impair performance than to favor it.

We motivated our keypoint candidate optimization module with the problem of handling symmetries in an object. As we can observe in Fig.~\ref{fig:ablations}, the optimization successfully avoids a collapse in the prediction, which is the reason for the lower performance of a simple nearest-neighbor-based selection [KNN]. The retrieval of the most similar shape from the labeled samples before optimization does not prove to be effective with this low number of shots [Retrieval].

Finally, we also evaluate our proposed method when using only one labeled sample [1-Shot]. While the performance is worse than with three labeled samples and has a higher variance, it still outperforms the previous state-of-the-art for keypoint detection with three given labeled samples by a large margin, thus demonstrating the strength of our framework. In the supplementary materials, we additionally give insights into the hyperparameter search for $\alpha, \beta$ and $\sigma$.

\mypara{Other Shape Descriptors} \label{sec:shape-descriptors}

\begin{figure}[htb]
    \centering
    \begin{tikzpicture}
        \begin{axis}[
            axis x line = none,
            axis y line = none,
            legend style = {at = {(0.01, 0.57)}, anchor = north west, draw = none, nodes={scale=0.48, transform shape}, fill opacity = 1.0, text opacity = 1},
            legend cell align={left},
        ]
            \addplot[opacity=0,forget plot] {0};
            \addlegendimage{/pgfplots/refstyle=descriptors-dino} \addlegendentry{DINO}
            \addlegendimage{/pgfplots/refstyle=descriptors-EffNet} \addlegendentry{EffNet}
            \addlegendimage{/pgfplots/refstyle=descriptors-CLIP} \addlegendentry{CLIP}
            \addlegendimage{/pgfplots/refstyle=descriptors-SAM} \addlegendentry{SAM}
            \addlegendimage{/pgfplots/refstyle=descriptors-hks} \addlegendentry{HKS}
            \addlegendimage{/pgfplots/refstyle=descriptors-wks} \addlegendentry{WKS}
            \addlegendimage{/pgfplots/refstyle=descriptors-shot} \addlegendentry{SHOT}
            \addlegendimage{/pgfplots/refstyle=descriptors-fskd} \addlegendentry{FSKD}
        \end{axis}
        \begin{axis}[
            width=.85\columnwidth,
            height=.58\columnwidth,
            xlabel={Distance Threshold},
            ylabel={IoU},
            xmin=-0.005, xmax=0.105,
            ymin=0, ymax=0.80,
            xtick={0, 0.02, 0.04, 0.06, 0.08, 0.1},
            ytick={0, 0.2, 0.4, 0.6},
            xticklabel=\pgfkeys{/pgf/number format/.cd,fixed,precision=2,zerofill}\pgfmathprintnumber{\tick},
            ylabel near ticks,
            ymajorgrids=true,
            grid style=dashed,
            every axis plot/.append style={line width=1.5pt}
            ]
            
            \addplot[color=plot0] table [x=Threshold, y=Ours, col sep=comma] {numbers/ablation-descriptors.csv}; \label{descriptors-dino}
            \addplot[name path=full_top, color=plot0, draw opacity=0] table [x=Threshold, y=Ours-Up, col sep=comma] {numbers/ablation-descriptors.csv};
            \addplot[name path=full_down, color=plot0, draw opacity=0] table [x=Threshold, y=Ours-Down, col sep=comma] {numbers/ablation-descriptors.csv};
            \addplot[plot0!50,fill opacity=0.5] fill between[of=full_top and full_down];
            
            \addplot[color=plot1] table [x=Threshold, y=HKS, col sep=comma] {numbers/ablation-descriptors.csv}; \label{descriptors-hks}
            \addplot[name path=hks_top, color=plot1, draw opacity=0] table [x=Threshold, y=HKS-Up, col sep=comma] {numbers/ablation-descriptors.csv};
            \addplot[name path=hks_down, color=plot1, draw opacity=0] table [x=Threshold, y=HKS-Down, col sep=comma] {numbers/ablation-descriptors.csv};
            \addplot[plot1!50,fill opacity=0.5] fill between[of=hks_top and hks_down];

            \addplot[color=plot2] table [x=Threshold, y=WKS, col sep=comma] {numbers/ablation-descriptors.csv}; \label{descriptors-wks}
            \addplot[name path=wks_top, color=plot2, draw opacity=0] table [x=Threshold, y=WKS-Up, col sep=comma] {numbers/ablation-descriptors.csv};
            \addplot[name path=wks_down, color=plot2, draw opacity=0] table [x=Threshold, y=WKS-Down, col sep=comma] {numbers/ablation-descriptors.csv};
            \addplot[plot2!50,fill opacity=0.5] fill between[of=wks_top and wks_down];

            \addplot[color=plot3] table [x=Threshold, y=SHOT, col sep=comma] {numbers/ablation-descriptors.csv}; \label{descriptors-shot}
            \addplot[name path=shot_top, color=plot3, draw opacity=0] table [x=Threshold, y=SHOT-Up, col sep=comma] {numbers/ablation-descriptors.csv};
            \addplot[name path=shot_down, color=plot3, draw opacity=0] table [x=Threshold, y=SHOT-Down, col sep=comma] {numbers/ablation-descriptors.csv};
            \addplot[plot3!50,fill opacity=0.5] fill between[of=shot_top and shot_down];

            \addplot[color=plot4] table [x=Threshold, y=CLIP, col sep=comma] {numbers/ablation-descriptors.csv}; \label{descriptors-CLIP}
            \addplot[name path=CLIP_top, color=plot4, draw opacity=0] table [x=Threshold, y=CLIP-Up, col sep=comma] {numbers/ablation-descriptors.csv};
            \addplot[name path=CLIP_down, color=plot4, draw opacity=0] table [x=Threshold, y=CLIP-Down, col sep=comma] {numbers/ablation-descriptors.csv};
            \addplot[plot4!50,fill opacity=0.5] fill between[of=CLIP_top and CLIP_down];

            \addplot[color=plot5] table [x=Threshold, y=SAM, col sep=comma] {numbers/ablation-descriptors.csv}; \label{descriptors-SAM}
            \addplot[name path=SAM_top, color=plot5, draw opacity=0] table [x=Threshold, y=SAM-Up, col sep=comma] {numbers/ablation-descriptors.csv};
            \addplot[name path=SAM_down, color=plot5, draw opacity=0] table [x=Threshold, y=SAM-Down, col sep=comma] {numbers/ablation-descriptors.csv};
            \addplot[plot5!50,fill opacity=0.5] fill between[of=SAM_top and SAM_down];

            \addplot[color=plot6] table [x=Threshold, y=EffNet, col sep=comma] {numbers/ablation-descriptors.csv}; \label{descriptors-EffNet}
            \addplot[name path=EffNet_top, color=plot6, draw opacity=0] table [x=Threshold, y=EffNet-Up, col sep=comma] {numbers/ablation-descriptors.csv};
            \addplot[name path=EffNet_down, color=plot6, draw opacity=0] table [x=Threshold, y=EffNet-Down, col sep=comma] {numbers/ablation-descriptors.csv};
            \addplot[plot6!50,fill opacity=0.5] fill between[of=EffNet_top and EffNet_down];
            
            \addplot[color=plot8, dashed] table [x=Threshold, y=FSKD, col sep=comma] {numbers/all.csv}; \label{descriptors-fskd}
        \end{axis}
    \end{tikzpicture}
    \caption{Experimental results when using different features with our proposed keypoint optimization module. We find that traditional shape descriptors cannot reach the performance of back-projected features of 2D foundation models, while other back-projected features cannot handle local geometry as well as DINO.}
    \label{fig:few-shot-others}
\end{figure}
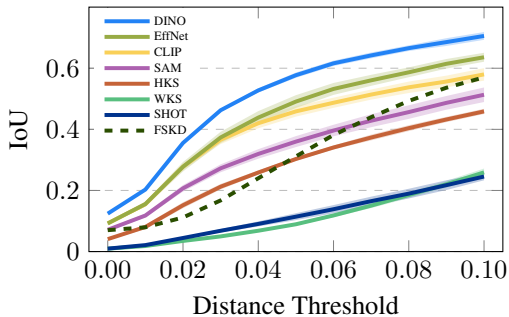

We want to evaluate the contribution of the back-projected features and compare our framework with the same keypoint candidate optimization but with other shape descriptors. We first compare against traditional, geometry-based shape descriptors HKS~\citep{sun2009concise}, WKS~\citep{aubry2011wave} and SHOT~\citep{salti2014shot}.
As these geometry-based descriptors are sensitive to low-quality meshes, we pre-process the given ShapeNet models to obtain clean watertight manifolds~\citep{huang2020manifoldplus}.
After optimization of hyperparameters, like the diffusion time for HKS and WKS, we report the results of using our keypoint optimization module together with the geometry-based descriptors in Fig.~\ref{fig:few-shot-others}.

In addition, we compare deep features back-projected from CLIP~\citep{radford2021learning} and SAM~\citep{kirillov2023segment}. We extract the features from the last layer of the respective image-encoding vision transformers, discarding subsequent layers that no longer provide significant spatial information. For additional comparison, we also extract features from the CNN EfficientNet~\citep{tan2019efficientnet}.
While all deep methods' features perform better than the traditional descriptors, they fall short of the features extracted from the DINO model. Our method with CLIP and EfficientNet features also outperforms the previous SOTA FSKD.

To our surprise, features extracted from the smaller and older EfficientNet outperform the CLIP and SAM models. We attribute this performance potentially to the fact that CLIP and SAM have specific properties (i.e., the joint image-text embedding for CLIP and the segmentation decoder for SAM) that may not necessarily be relevant in our setting. However, a more thorough investigation is necessary to fully understand the exact properties of the different features.

While the HKS performs the best out of the three traditional shape descriptors, the results are only as good as simple farthest-point sampling in our experiments.
Since such hand-crafted methods are not aware of semantic information, they struggle to provide similar features for similar points on shapes of the same object class that have a different geometry. To validate this conjecture, we compare the features computed across multiple shapes of the same class in the supplementary materials.

\subsection{Part Segmentation Transfer}

As an additional validation of our results, we investigate the performance of back-projected features for transferring part segmentation labels between a pair of shapes. Using the back-projected DINO features and a simple nearest-neighbor-based classification, we obtain an average IoU of 71.0\% on the ShapeNet part dataset \citep{yi2016scalable}, improving by nearly 2\% over the previous state-of-the-art NCP~\citep{attaiki2022ncp} (69.2\% IoU). Further details and results with features back-projected from other 2D models can be found in the supplementary materials.

\section{Conclusions}

We presented B2-3D, a novel method for few-shot keypoint detection on 3D meshes. Our method consists of back-projecting features from powerful pre-trained 2D vision encoders to the 3D shape, which carry strong semantic and geometric information, as we were able to show in a comprehensive feature analysis. In order to transfer keypoints between shapes, we match the observed keypoint distributions on the shape with a simple yet effective optimization strategy that is agnostic of the specific shape descriptor used. We demonstrated the effectiveness of our formulation by achieving state-of-the-art performance on the KeypointNet dataset by a large margin in combination with back-projected DINO features, even outperforming the previous SOTA with back-projected CNN features. We further achieve a new state-of-the-art for the task of part segmentation transfer.

While the proposed feature extraction is learning-free, the computation cost of features for multiple views around the object is slightly higher than with pure 3D-based methods. A remaining difficulty is filtering out non-existent keypoints on new shapes and detecting unseen keypoint classes. While we propose to solve the first problem using shape retrieval, the diversity and representativeness of the given labeled shapes are still essential for these problems.

Back-projected features can serve as a powerful prior for various other shape analysis tasks where pure geometric methods currently still fail, which is an exciting avenue for future research. The back-projection of different 2D features for such downstream tasks can also serve as a powerful testbed for comparing the quality of learned features, especially on photo-realistic datasets~\citep{bordes2023pug}.

\paragraph*{Acknowledgements.}
Thomas Wimmer is supported by the Konrad Zuse School of Excellence in Learning and Intelligent Systems (ELIZA) through the DAAD programme Konrad Zuse Schools of Excellence in Artificial Intelligence, sponsored by the German Federal Ministry of Education and Research.
Parts of this work were supported by the ERC Starting Grant 758800 (EXPROTEA), ERC Consolidator Grant 101087347 (VEGA), ANR AI Chair AIGRETTE, and gifts from Adobe Inc. and Ansys Inc.

{
    \small
    \bibliographystyle{ieeenat_fullname}
    \bibliography{main}
}

\clearpage
\setcounter{page}{1}
\maketitlesupplementary

\section{Implementation Details}

\begin{figure}
    \centering
    \includegraphics[width=\columnwidth]{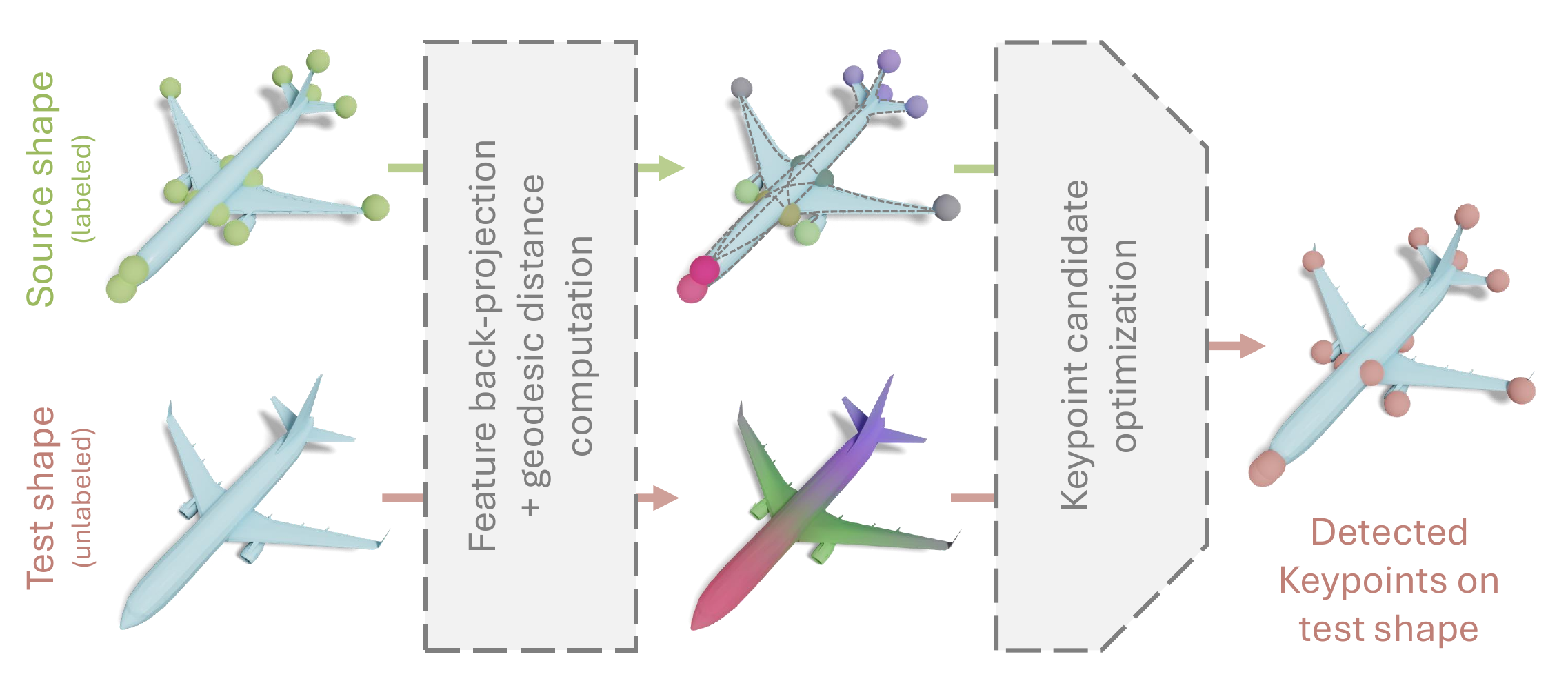}
    \caption{Pipeline of B2-3D. Given one or a several labeled source shapes and an unlabeled \textit{test shape}, we first back-project features for ground-truth keypoints on the source shapes and candidate locations sampled from the surface of the test shape. In addition, we compute the pairwise geodesic distances between the keypoints. With this information at hand we then employ our optimization module to detect the keypoints on the test shape. Intuitively, our optimization uses the back-projected features as \textit{first order} similarity between labeled keypoints and candidate locations, and uses pairwise  geodesic distance information as  \textit{second order} regularization.}
    \label{fig:pipeline}
\end{figure}

The pipeline of our method is visualized in Fig.~\ref{fig:pipeline}. The features and the pairwise geodesic distances of ground-truth keypoints on the few-shot samples can be computed in advance.
We use PyTorch3D~\citep{ravi2020pytorch3d} for the rendering of shapes, and process the rendered views with the pre-trained vision models. The runtime of the feature extraction thus depends on various factors, like the number of views, the complexity of the rendered mesh, the amount of points for which we extract features and the 2D feature extractor that is used. The feature computation normally takes between 10 to 20 seconds for one shape in our experiments on one NVIDIA A40 GPU.

As the keypoint candidate optimization is non-linear, we use PyTorch with gradient descent to solve the optimization problem. We initialize the matrix $S$ (see Sec.~\ref{sec:optimization}) with random values from a normal distribution and apply a softmax per row to ensure the right-stochastic character of the matrix $S$ at every optimization step. Using a GPU, the optimization process (5000 steps) can be completed in about 10 seconds. We make our code available under the following URL: \url{https://github.com/wimmerth/back-to-3d-few-shot-keypoints}.

\subsection{Keypoint optimization hyperparameters}

To further steer the optimization toward the selection of one clear correspondence per keypoint, instead of possibly averaging over multiple candidates and their features in the optimization, we define an optional selection reward
\begin{equation}
    R_{\text{selection}} = \sum_j (\max_i \hat{S}_{ij} - \frac{1}{n} \sum_i \hat{S}_{ij}),
\end{equation}
and formulate an extended objective function as
\begin{equation}\label{eq:objective-function}
    L = L_{\text{feature}} + \alpha L_{\text{distance}} - \beta R_{\text{selection}}
\end{equation}
with two weighting parameters $\alpha,\beta$.

\begin{figure}[htb]
    \centering
    \hspace*{\fill}%
    \begin{subfigure}[b]{.45\columnwidth}
        \begin{tikzpicture}
        \begin{axis}[
            width=\columnwidth,
            height=\columnwidth,
            legend style = {at = {(0.98501, 0.01)}, anchor = south east, draw = none, nodes={scale=0.6, transform shape}, fill opacity = 0.0, text opacity = 1},
            legend cell align={left},
            xlabel={Distance Threshold},
            ylabel={IoU},
            xmin=-0.005, xmax=0.105,
            ymin=0, ymax=0.75,
            xtick={0.00, 0.05, 0.10},
            ytick={0, 0.2, 0.4, 0.6},
            xticklabel=\pgfkeys{/pgf/number format/.cd,fixed,precision=2}\pgfmathprintnumber{\tick},
            ymajorgrids=true,
            grid style=dashed,
            every axis plot/.append style={line width=1.pt},
            legend image code/.code={
                \draw[mark repeat=2,mark phase=2]
                plot coordinates {
                (0cm,0cm)
                (0.15cm,0cm)        
                (0.3cm,0cm)         
                };%
            }
            ylabel near ticks,
            ]
            \addlegendimage{empty legend}; \addlegendentry{\hspace{-.5cm}$\bm{\alpha}$};
            \addplot[color=alpha0] table [x=X, y=1, col sep=comma] {numbers/alpha.csv}; \addlegendentry{1};
            \addplot[color=alpha1] table [x=X, y=2, col sep=comma] {numbers/alpha.csv}; \addlegendentry{2};
            \addplot[color=alpha2] table [x=X, y=3, col sep=comma] {numbers/alpha.csv}; \addlegendentry{3};
            \addplot[color=alpha3] table [x=X, y=4, col sep=comma] {numbers/alpha.csv}; \addlegendentry{4};
            \addplot[color=alpha4] table [x=X, y=5, col sep=comma] {numbers/alpha.csv}; \addlegendentry{5};
        \end{axis}
    \end{tikzpicture}
    \end{subfigure}
    \hspace*{\fill}%
    \begin{subfigure}[b]{.45\columnwidth}
        \begin{tikzpicture}
        \begin{axis}[
            width=\columnwidth,
            height=\columnwidth,
            legend style = {at = {(0.98501, 0.01)}, anchor = south east, draw = none, nodes={scale=0.6, transform shape}, fill opacity = 0.0, text opacity = 1},
            legend cell align={left},
            legend image code/.code={
                \draw[mark repeat=2,mark phase=2]
                plot coordinates {
                (0cm,0cm)
                (0.15cm,0cm)        
                (0.3cm,0cm)         
                };%
            },
            xlabel={Distance Threshold},
            ylabel={IoU},
            xmin=-0.005, xmax=0.105,
            ymin=0, ymax=0.75,
            xtick={0.00, 0.05, 0.10},
            ytick={0, 0.2, 0.4, 0.6},
            xticklabel=\pgfkeys{/pgf/number format/.cd,fixed,precision=2}\pgfmathprintnumber{\tick},
            ymajorgrids=true,
            grid style=dashed,
            every axis plot/.append style={line width=1.5pt},
            ylabel near ticks,
            ]
            \addlegendimage{empty legend}; \addlegendentry{\hspace{-.5cm}$\bm{\beta}$};
            \addplot[color=beta2] table [x=X, y=0, col sep=comma] {numbers/beta.csv}; \addlegendentry{0};
            \addplot[color=beta1] table [x=X, y=1, col sep=comma] {numbers/beta.csv}; \addlegendentry{1};
            \addplot[color=beta0] table [x=X, y=2, col sep=comma] {numbers/beta.csv}; \addlegendentry{2};
        \end{axis}
    \end{tikzpicture}
    \end{subfigure}
    \hspace*{\fill}%
    \caption{Influence of the weighting terms $\alpha, \beta$ in the optimization objective on the keypoint detection results. The best weights with DINO features were empirically found to be $\alpha=4$ and $\beta=0$, thus effectively removing the selection reward from the objective.}
    \label{fig:few-shot-ablation-alpha-beta}
\end{figure}
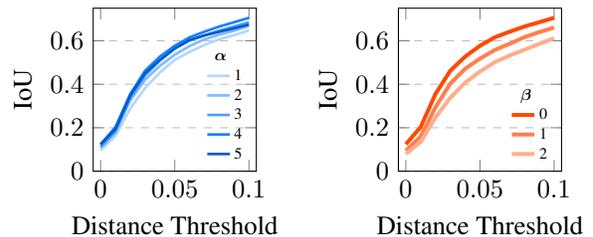

\begin{itemize}
    \item The weighting-parameter $\alpha$ is dependent on the feature dimensionality and magnitude. We find that setting $\alpha=4$ works well with back-projected DINO features (see Fig.~\ref{fig:few-shot-ablation-alpha-beta}), but we also adjust this parameter for the various other features used in our experiments.
    \item In our experiments, we find that setting $\beta = 0$ in the optimization and thus not taking the selection reward into account works the best (Fig.~\ref{fig:few-shot-ablation-alpha-beta}), as the optimization is less sensitive to the random initialization of the selection matrix $S$ without it.
    \item The choice of $\sigma$ for Gaussian re-weighting of features (Eq.~\ref{eq:gaussian-reweighting}) depends on the quality and scale of the given mesh. In the best case, if we work with clean meshes, we do not need to apply re-weighting since all points of the shape will be (recognized as) visible from some viewpoint. However, if this is not the case, as with some ShapeNet meshes, we find that setting $\gamma\in[0.001,0.005]$ works quite well for shapes normalized to a unit box scale.
\end{itemize}

\subsection{Hungarian method baseline}
An alternative baseline to nearest-neighbor matching in the feature space is using the Hungarian method to solve the linear assignment problem with the similarity of candidate features to keypoint features as costs. We show the results using a variant, the Jonker-Volgenant algorithm, in Tab.~\ref{tab:hungarian-baseline}.
While this method avoids the collapse observed with simple nearest-neighbor matching in some cases, it is not aware of the spatial relation between the detected keypoints and using our proposed optimization module instead improves over the results by an average of 37\%, thus strongly supporting our design choice.

\begin{table}[ht!]
\caption{Comparison of IoU scores with varying distance thresholds against proposed baseline using Hungarian matching.}
\label{tab:hungarian-baseline}
\resizebox{\columnwidth}{!}{
\begin{tabular}{l|lllllllllll}
Dist. Threshold & 0.00     & 0.01  & 0.02  & 0.03  & 0.04  & 0.05  & 0.06  & 0.07  & 0.08  & 0.09  & 0.10   \\ \hline
B2-3D (ours) & 0.12 & 0.20 & 0.36 & 0.46 & 0.53 & 0.58 & 0.62 & 0.64 & 0.67 & 0.69 & 0.71 \\
Hungarian & 0.08 & 0.13 & 0.23 & 0.31 & 0.37 & 0.42 & 0.47 & 0.50 & 0.53 & 0.55 & 0.58
\end{tabular}
}
\end{table}

\section{Results on real-world scans}

In addition to experiments on the KeypointNet dataset, we qualitatively evaluate our method on real-world scans from the Objaverse dataset~\citep{deitke2023objaverse}.
As there are no ground-truth keypoint annotations given, we manually annotate keypoints on a few cars and apply B2-3D to a few unlabeled cars. 
We observe that, contrary to the experiments on the KeypointNet dataset, using texture information of the models slightly improves the results which can be explained with the higher quality of the texture compared to ShapeNet meshes.

\begin{figure}[ht]
    \centering
    \includegraphics[width=\columnwidth]{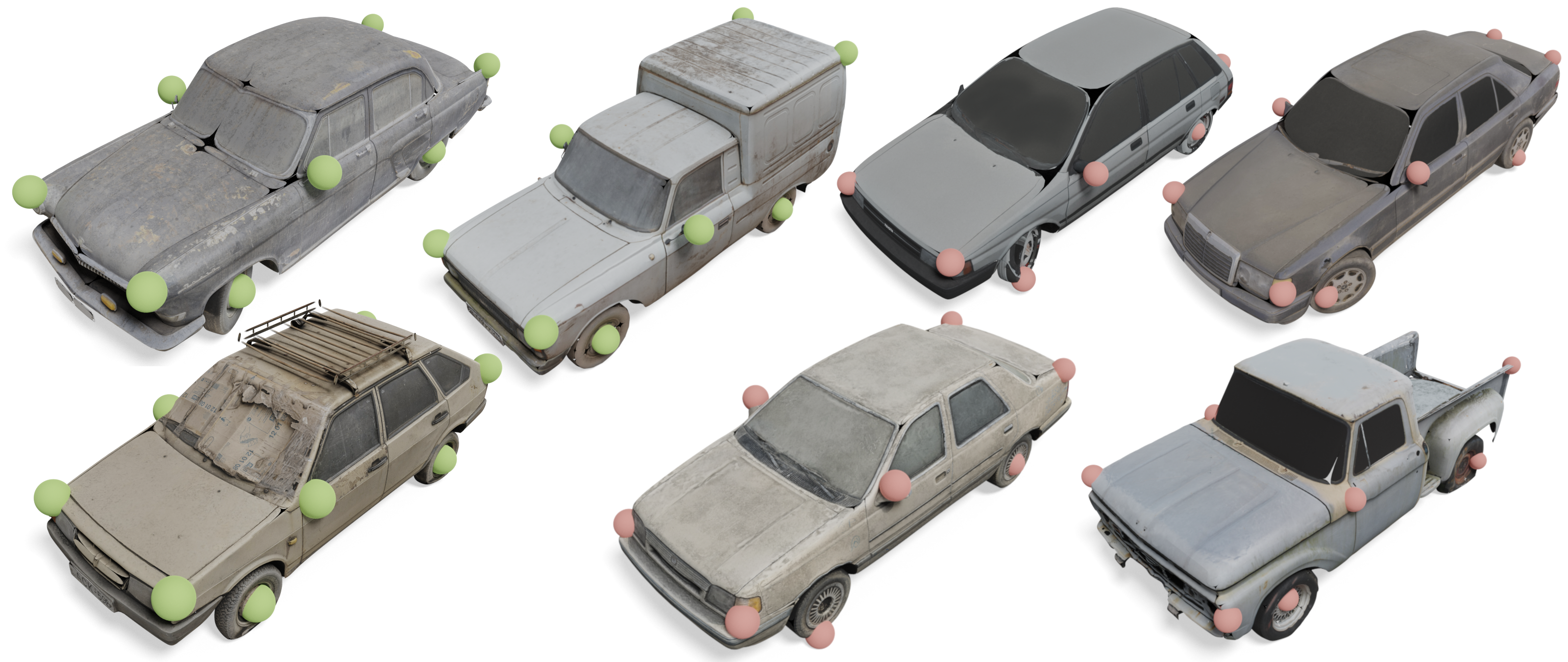}
    \caption{Few-shot keypoint detection (red) on real-world scans, given manually annotated source shapes (green).}
    \label{fig:enter-label}
\end{figure}

\section{Additional feature analysis and properties}

\subsection{Uptake of local geometry changes}

\begin{figure}[htb]
    \centering
    \includegraphics[width=\columnwidth]{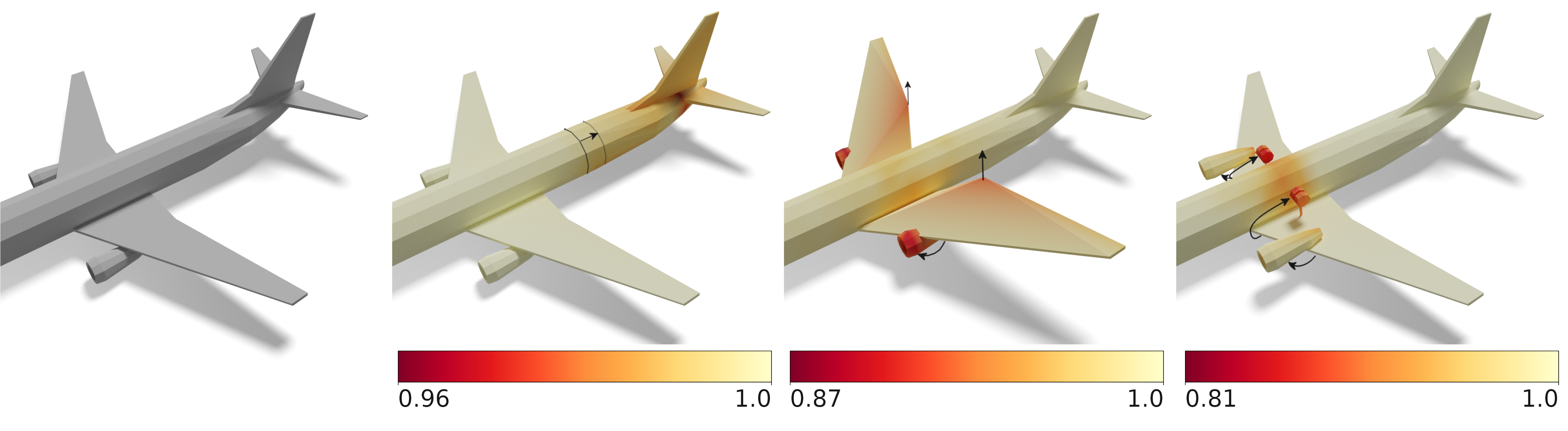}
    \caption{Change of the computed features (measured in cosine similarity) when applying small modifications (indicated with black arrows) to the original shape (left). The back-projected DINO features seem to generally react well by changing in the affected areas while staying the same in unaffected areas.}
    \label{fig:deformation-anaysis}
\end{figure}

We investigate the reaction of the back-projected features to small modifications of the shape.
We expect them to slightly change for the affected regions while remaining similar for non-affected regions. The visual results of this experiment can be seen in Fig.~\ref{fig:deformation-anaysis}.

If we slightly stretch the rear part of the aircraft body, we notice that the features of the aircraft change slightly from the point where we stretch it to the tail. 
This is interesting to observe because only the fuselage was changed, while the stabilizers in the rear remained untouched. We suspect that the change in the relative sizes of the different parts influences these small changes in the features.

Moving the landing gear on the top of the aircraft changes not only the features of the landing gear, but also those of the part of the aircraft to which it is attached. While it could be argued that the coarse patch size of the extracted features affects this change by "leaking" onto the area behind the landing gear, features for other regions that are close to the landing gear did not change drastically. This leads us to conclude that the features back-projected to the modified shape also capture the change in semantics for the affected part of the aircraft.

\begin{table*}[htb!]
\centering
\caption{Back-projected features are a strong backbone for part segmentation transfer. Label transfer results measured with average IoU.}
\label{tab:label-transfer-results}
\resizebox{\textwidth}{!}{
\begin{tabular}{clccccccccccccccc|c}
\hline & & & & & & & & & & & & & & & & &\\[-1.9ex]
 & & \textit{\textbf{pla.}} & \textit{\textbf{bag}} & \textit{\textbf{cap}} & \textit{\textbf{cha.}} & \textit{\textbf{ear.}} & \textit{\textbf{gui.}} & \textit{\textbf{kni.}} & \textit{\textbf{lam.}} & \textit{\textbf{lap.}} & \textit{\textbf{bik.}} & \textit{\textbf{mug}} & \textit{\textbf{pis.}} & \textit{\textbf{roc.}} & \textit{\textbf{ska.}} & \textit{\textbf{tab.}} & \textit{\textbf{avg.}} \\ [1pt] \hline & & & & & & & & & & & & & & & &\\[-1.7ex]
\parbox[t]{3mm}{\citep{liu2020learning}} & IDC & 60.1 & 56.2 & 59.7 & 72.2 & 45.3 & 81.5 & 66.4 & 42.6 & 88.5 & 40.5 & 87.5 & 66.4 & 37.2 & 50.7 & 70.4 & 61.7 \\
\parbox[t]{3mm}{\citep{cheng2021learning}}& CPAE & 61.3 & 59.3 & 61.6 & 72.6 & 55.5 & 78.9 & 71.3 & 53.2 & 89.9 & 55.4 & 86.5 & 66.2 & 40.2 & 61.8 & 72.5 & 65.8 \\
\parbox[t]{3mm}{\citep{attaiki2022ncp}}& NCP & 63.7 & 66.7 & 68.7 & \textbf{80.2} & 59 & 78.8 & 72.5 & \textbf{61.9} & \textbf{91.4} & 57.2 & 89.5 & 61.4 & 44.2 & 63.6 & \textbf{79.2} & 69.2 \\[1pt] \hline & & & & & & & & & & & & & & & & &\\[-1.7ex]
\parbox[t]{1mm}{\multirow{4}{*}{\rotatebox[origin=c]{90}{Ours}}} & DINO & \textbf{64.8} & \textbf{75.1} & 67.5 & 72.1 & \textbf{77.3} & \textbf{83.6} & 71.7 & 57.9 & 89.6 & \textbf{64.3} & \textbf{91.0} & \textbf{73.2} & \textbf{47.9} & \textbf{65.8} & 63.1 & \textbf{71.0} \\
& CLIP & 59.4 & 66.9 & \textbf{73.4} & 62.9 & 70.9 & 76.4 & 68.1 & 52.8 & 84.1 & 57.3 & 86.4 & 66.1 & 43.4 & 63.6 & 62.2 & 66.3 \\
& EffNet & 56.6 & 69.5 & 63.2 & 64.4 & 72.1 & 81.6 & 70.8 & 50.6 & 88.5 & 59.5 & 82.6 & 63.5 & 44.9 & 60.2 & 64.2 & 66.1 \\
& SAM & 36.8 & 65.3 & 57.1 & 59.7 & 56.6 & 79.0 & \textbf{72.8} & 51.8 & 75.8 & 43.6 & 49.0 & 56.6 & 32.4 & 46.8 & 53.8 & 55.8 \\ [1pt] \hline
\end{tabular}
}
\end{table*}

\subsection{Semantic stability with varying shapes}

\begin{figure}[htb]
    \centering
    \begin{subfigure}[b]{\columnwidth}
        \centering
        \includegraphics[width=\columnwidth]{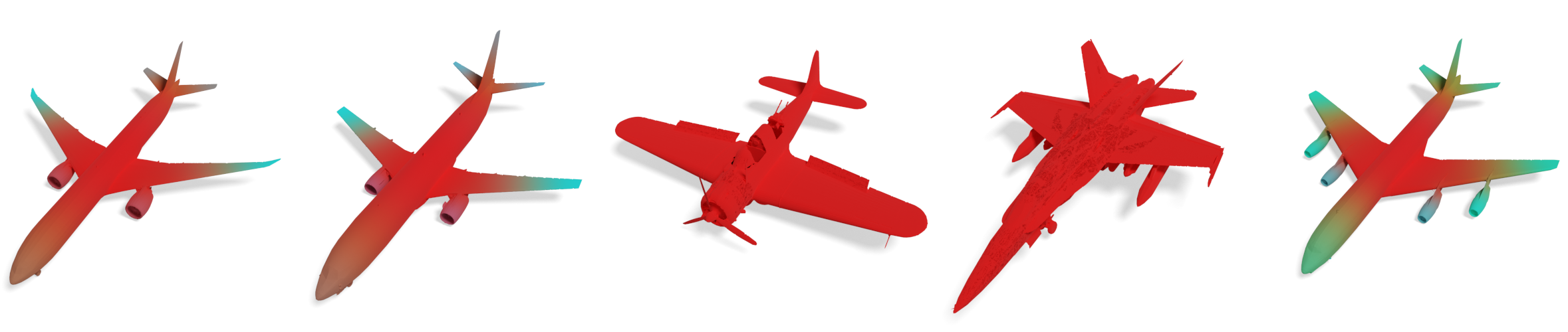}
        \caption{HKS features on different shapes of the same class.}
    \end{subfigure}
    \begin{subfigure}[b]{\columnwidth}
        \centering
        \includegraphics[width=\columnwidth]{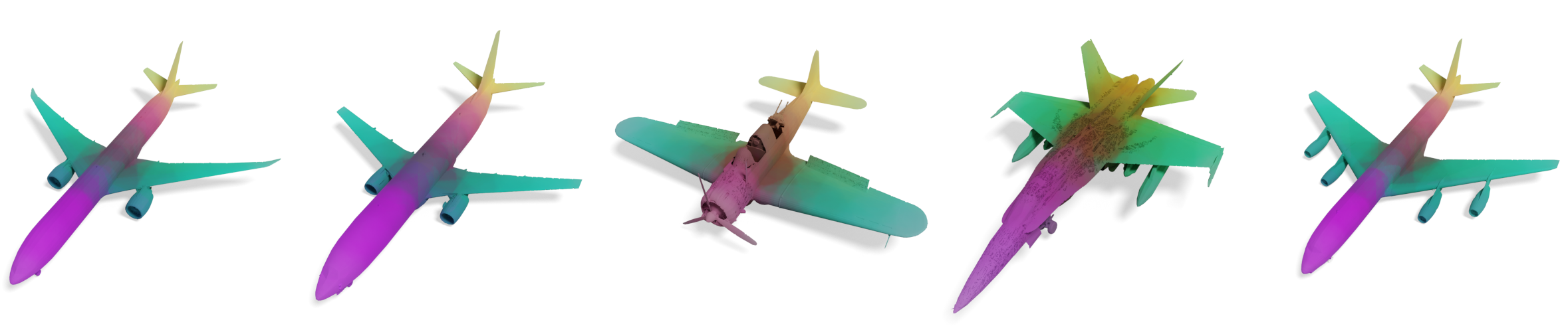}
        \caption{Back-projected DINO features on different shapes of the same class.}
    \end{subfigure}
    \caption{The back-projected features produce more consistent and distinctive features for similar points on different shapes.}
    \label{fig:few-shot-others-explanation}
\end{figure}

In our experiments with axiomatic shape descriptors, we hypothesize that pure geometry-based descriptors are not consistent and informative enough to provide similar features for similar points on shapes of the same object class that have a different geometry.
To test this conjecture, we conduct another small qualitative experiment.
We extract features for each shape and apply a PCA to all points together for visualization, as described in Section~\ref{sec:semantic-analysis}. In order to have a balanced PCA computation between the meshes with different numbers of vertices, we sample 2048 points from each shape's surface which we use to fit the PCA. For visualization, we then apply the fitted PCA to the feature computed for each mesh vertex to obtain the colored shapes shown in Fig.~\ref{fig:few-shot-others-explanation}.

We would want shapes to be colored approximately the same, as similar parts over different shapes should also have similar features.
The results (Fig.~\ref{fig:few-shot-others-explanation}) show that this is far more the case for the back-projected DINO features when compared to the HKS features.
It is remarkable that the features are consistent even for shapes that are fairly different, such as different airplane types.

\section{Part Segmentation Transfer}

To further validate the strength and generalizability of the back-projected features, we evaluate them on the task of part label transfer. In our experiments, we follow the setup of \citet{cheng2021learning}, where the goal is to transfer part segmentation labels from one shape to another using the labels from the ShapeNet part dataset~\citep{yi2016scalable}.

In our experiments, we back-project the features onto the 3D shape, as described in Section~\ref{sec:feature-extraction} of the main paper, and perform a $k$-nearest neighbor classification in the feature space, querying the points on the new shape and retrieving the best matching label for each point.

Using DINO features, we outperform the previous state-of-the-art methods in 9 of the 15 categories, with an increase of the average IoU over all classes by almost three percent (see Tab.~\ref{tab:label-transfer-results}). The semantic information enables gains of up to 18.3\% IoU over previous methods for selected classes.
While these results underline the strong performance of back-projected features, we want to emphasize the simplicity of the used approach: The nearest-neighbor-based classification in feature space is not informed by the spatial connectivity of points, etc., and the used approach is much faster than the previously proposed methods that require additional optimization~\citep{liu2020learning, attaiki2022ncp}.

Our experiments on this part segmentation transfer further highlight the superiority of DINO features, as the back-projected DINO features outperform other back-projected features on this task. As with keypoint detection, CLIP and EfficientNet features provide fairly similar results. Even though the SAM model was trained as a foundation model for segmentation tasks, the extracted features seem to not be able to provide the necessary distinctiveness between different object parts. We suggest investigating the reasons for this behavior further, but it could possibly be explained by the absence of SAM's powerful decoder that further processes the extracted features and is possibly responsible for the performance increase in the 2D setting.

\end{document}